\documentclass{article}

\usepackage{preprint_layout,times}

\usepackage{amsmath,amsfonts,bm}

\def\eqref#1{equation~\ref{#1}}

\def\1{\bm{1}}

\DeclareMathAlphabet{\mathsfit}{\encodingdefault}{\sfdefault}{m}{sl}
\SetMathAlphabet{\mathsfit}{bold}{\encodingdefault}{\sfdefault}{bx}{n}

\usepackage{amsmath,amssymb}
\usepackage{booktabs}
\usepackage{algorithm}
\usepackage{graphicx}
\usepackage{microtype}
\usepackage{hyperref}
\hypersetup{
  hidelinks,
  pdftitle={Grammar-Weighted Futures: Modeling Human Expectation with Open Goals},
  pdfauthor={Xiaoxuan Wang, Vincent K.M. Cheung}
}
\usepackage{url}

\newcommand{\conceptplot}{\includegraphics{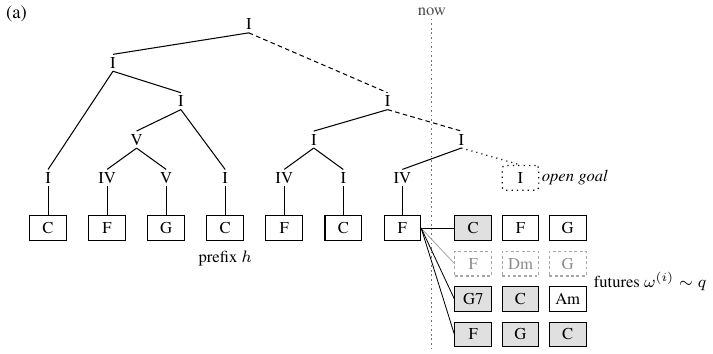}}
\newcommand{\methodoverview}{\includegraphics{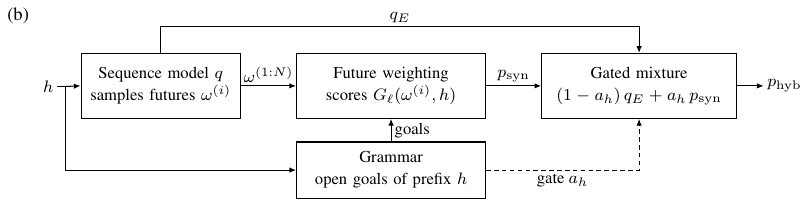}}

\newcommand{\languagehorizonplot}{\includegraphics{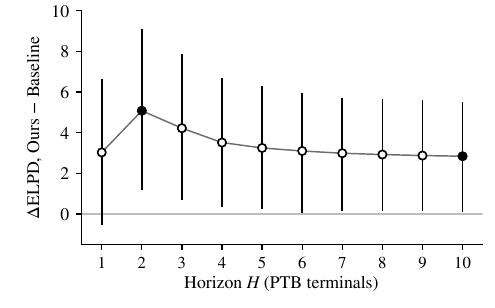}}
\newcommand{\workedexampleplot}{\includegraphics[width=\linewidth]{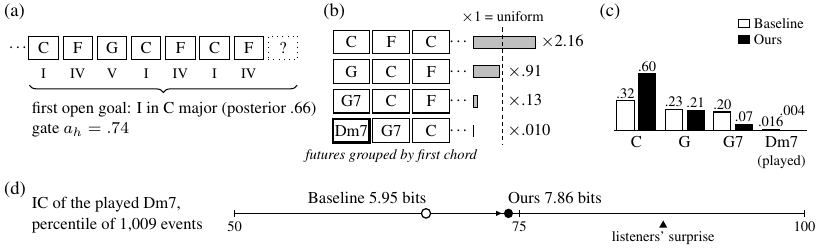}}

\title{Grammar-Weighted Futures: Modeling Human Expectation with Open Goals}

\author{%
  {\large\textbf{Xiaoxuan Wang}\textsuperscript{1}\qquad
  \textbf{Vincent K.M.\ Cheung}\textsuperscript{2}}\\[0.7em]
  \textsuperscript{1}École polytechnique fédérale de Lausanne\\
  \href{mailto:xiaoxuan.wang@epfl.ch}{\texttt{xiaoxuan.wang@epfl.ch}}\\[0.45em]
  \textsuperscript{2}Sony Computer Science Laboratories\\
  \href{mailto:cheung@csl.sony.co.jp}{\texttt{cheung@csl.sony.co.jp}}%
}

\begin{document}

\maketitle

\begin{abstract}
Behavioral and neural evidence suggests that human expectations in language and
music are not driven by surface statistics alone: readers and listeners also
anticipate how open goals will be completed, as a subject awaits its verb and a
chord progression awaits its return to the home chord. We present
Grammar-Weighted Futures, a model in which such anticipation constrains the
probability of the next event. A fixed sequence model, which predicts
events well without exposing open goals, samples continuations; a probabilistic
grammar, which represents open goals explicitly but predicts events poorly,
scores how well each continuation completes the goals of the observed prefix;
and these scores reweight the sequence model's prediction of the next event. Tested
against listeners' surprise ratings for chord progressions with Transformer,
LSTM and variable-order sequence models, grammar weighting improves prediction
for all three models on ratings held out from the response regression. In language, an exploratory implementation improves the
prediction of word-predictability ratings over GPT-2 alone. These results show that grammar weighting improves the prediction of human
expectation ratings. The added value of sampled futures is model-dependent.
\end{abstract}

\section{Introduction}

Theories of expectation in language and music commonly quantify how expected an event is by its probability given the preceding context \citep{hale2001earley,levy2008expectation,pearceWiggins2012auditory}. Yet readers and listeners also anticipate how structure that has begun will be completed. We call a unit of structure that the context predicts but that has not yet begun an \emph{open goal}. Readers who encounter \emph{either} anticipate a later \emph{or} \citep{staubClifton2006either}; holding predicted syntactic heads in memory incurs processing cost \citep{gibson1998locality,chenEtAl2005storage}; and cortical activity tracks the number of open constituents and drops when they close \citep{nelsonEtAl2017phrase}. Music has comparable structure. A chord is several notes sounded together, and a chord progression is a sequence of chords; each piece has a key whose home chord (the \textit{tonic}, written I) sounds stable, so a progression sets up the open goal of returning to I. Listeners' brain responses distinguish progressions that return to their initial key from those that do not \citep{koelschEtAl2013hierarchical}, and hierarchical models predict when listeners expect a progression to end \citep{herffEtAl2021completion}.

An open goal need not determine the next event: a progression may linger on other chords before it returns home. Representing open goals therefore need not improve next-event prediction, the criterion by which computational models of expectation are usually trained and evaluated. We ask whether an explicit account of open goals and their possible completions nonetheless improves the modeling of human expectation. Our hypothesis is that expectations about how open goals will be completed constrain the probabilities of upcoming events, even where these expectations do not raise the probability of what occurs.

\begin{figure}[t]
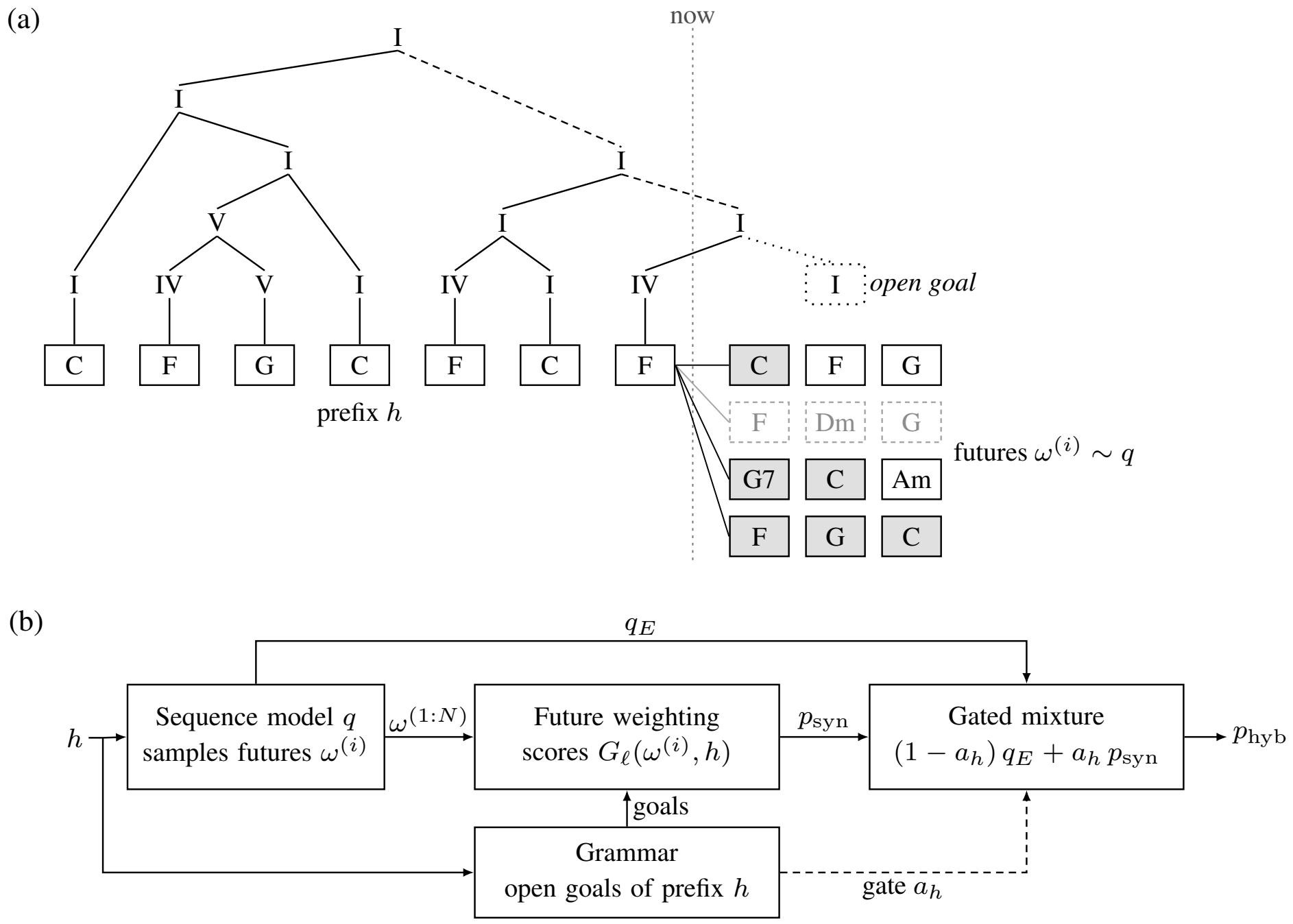

\centering
\resizebox{0.9\linewidth}{!}{\begin{tabular}{@{}l@{}}\conceptplot\\[6pt]\methodoverview\end{tabular}}
\caption{\textbf{Grammar-Weighted Futures.} (a) Illustrative schematic. Roman numerals mark each chord's role in the key; I is the home chord, C. Solid branches mark completed constituents, dashed ones constituents still growing, and the dotted node is an open goal, a predicted return to I. Sampled futures that reach C complete this goal (shaded) and score high; the grey one does not. (b) Computation; Section~\ref{sec:method} defines the symbols. The sequence model $q$ samples futures of the prefix $h$, and the grammar infers the prefix's open goals. Each future is scored by how well it completes these goals ($G_\ell$, at each completion length $\ell$). Adding up the normalized scores of futures that begin with the same event gives the grammar-weighted next-event distribution $p_{\mathrm{syn}}$. A gate $a_h$ computed from the prefix analysis (dashed) mixes $p_{\mathrm{syn}}$ with the sequence model's unchanged prediction $q_E$ (top line). The mixture's surprisal, $-\log_2 p_{\mathrm{hyb}}$, predicts human responses in a separate model. Next-event probabilities are computed without using ratings as inputs. Appendix~\ref{app:figure-example} gives a computed example.}
\label{fig:method}
\end{figure}

Two modeling traditions address distinct parts of this question. Sequence models learn the statistics of observed sequences. From variable-order Markov models such as IDyOM \citep{pearceWiggins2006expectation,pearceWiggins2012auditory} to neural networks, they predict the next event well; their predictions can reflect internally encoded structure, but they do not expose whether an open goal mediates a prediction. Generative grammars represent open goals explicitly \citep{lerdahlJackendoff1983generative,rohrmeier2011generative}, and prefix parsers give a grammar's own next-event probability by averaging over the ways the grammar can continue the prefix \citep{hale2001earley,levy2008expectation,stolcke1995efficient}, but grammars alone are weak sequence predictors. Rather than comparing the two \citep{frankBod2011insensitivity,fossumLevy2012sequential}, we combine them: the sequence model supplies fluent continuations, and the grammar supplies the criterion by which they are weighed. In music, both components receive the same chord history and the song's key. The grammar infers how chords group together and where they temporarily establish a different home chord, without phrase or section annotations.

Our model, \emph{Grammar-Weighted Futures}, makes this division of labor explicit (Figure~\ref{fig:method}): a fixed sequence model samples possible futures, a probabilistic grammar scores how well each future completes the open goals of the prefix, and the scores are turned into a next-event distribution that a gate mixes with the sequence model's own prediction. Under exact averaging, assigning the same positive score to every future leaves the sequence model's prediction unchanged. The hypothesis is stated at the computational level: we do not assume that people draw Monte Carlo samples.

We test three claims: H1 is the main claim, and H2 and H3 ask what produces the gain. H1: weighting futures by the grammar improves the prediction of human expectation over the sequence model alone; in music we also compare the model with a static preference for stable chords \citep{bharuchaKrumhansl1983stability,lerdahl2001tonal}. H2: the gain depends on assigning the grammar's scores to the corresponding futures, so neither shuffling the scores among the futures nor equally sharpening the sequence model's prediction without the grammar should reproduce it. H3: the sequence model's own sampled continuations add predictive value beyond a correction using the grammar's next-event probability, which already integrates possible goal completions.

We test these claims in two higher-level cognitive domains, music and language. In music, each chord maps to one grammar terminal, and listeners rated their surprise chord by chord \citep{cheungEtAl2024expectancy}; we test all claims with Transformer, LSTM and IDyOM sequence models and a grammar that describes each chord by its role in the local key \citep{harasimEtAl2018parsing}. In language, an exploratory implementation tests H1: combining GPT-2 \citep{radfordEtAl2019gpt2} with a treebank grammar predicts word-predictability ratings \citep{deVardaEtAl2024predictability} better than GPT-2 alone, with an uncertain increment beyond incremental alternative sampling (IAS; \citealp{giulianelli2026ias}).

Our contributions are twofold:
\begin{enumerate}
\item \emph{Grammar-Weighted Futures}, a way to add a probabilistic grammar's
knowledge of open goals to a trained sequence model without retraining it. The
grammar reweights the model's sampled continuations by how well they complete
open goals.
\item Evidence that grammar weighting improves models of human expectation without
improving event prediction. Grammar weighting better predicts
listeners' surprise ratings for three sequence models without making the chords
that occur more probable.
\end{enumerate}

\section{Related Work}

\paragraph{Surprisal and human expectation.}
Surprisal, or information content (IC), is the standard link between predictive models and human expectation in music \citep{pearceWiggins2012auditory} and in language, where it predicts reading times across eleven languages \citep{wilcoxEtAl2023surprisal} and N400 amplitudes \citep{michaelovEtAl2024strong}. Better next-event prediction does not guarantee a better account of behavior. Psychometric fit first improved with language-model quality \citep{goodkindBicknell2018surprisal}, but not in Japanese \citep{kuribayashiEtAl2021lower} and not for larger pretrained Transformers \citep{ohSchuler2023poorer}. These findings compare different models. We hold the trained sequence model fixed and compare different ways of weighting its continuation distribution.

\paragraph{Statistical and structural expectation.}
In music, statistical models such as IDyOM learn sequential regularities \citep{pearceWiggins2006expectation} and predict listeners' surprise at chords \citep{cheungEtAl2024expectancy}. Schematic knowledge of keys offers another source of expectation \citep{huron2006sweet}: listeners rank the notes of a key by how stable they sound \citep{krumhanslKessler1982tracing} and process chords related to preceding chords faster \citep{bharuchaStoeckig1986priming,bharucha1987music}, neither of which requires an open goal \citep{tillmannEtAl2000implicit}. Generative grammars of music make nested dependencies between chords explicit \citep{rohrmeier2011generative,harasimEtAl2018parsing}, hierarchical models of tonal tension fit listeners' tension ratings \citep{lerdahl2001tonal,lerdahlKrumhansl2007tension}, and the number of open dependencies predicts theta-band MEG activity \citep{herffEtAl2024theta}. In language, whether sequential models fit reading better than hierarchical ones depends on lexical controls \citep{frankBod2011insensitivity,fossumLevy2012sequential}, recurrent neural network grammars fit EEG and Japanese reading times better than LSTMs \citep{haleEtAl2018finding,yoshidaEtAl2021leftcorner}, and incremental parsers separate lexical from syntactic surprisal \citep{roarkEtAl2009lexical}. Open goals match predicted categories of \citet{gibson1998locality}, predicted nodes of \citet{demberg2013pltag} and open implications of \citet{rohrmeier2013expectancy}.

\paragraph{Sampled futures and reweighting.}
Anticipation has been measured as a grammar's uncertainty about the rest of the sentence \citep{hale2006uncertainty,linzenJaeger2016uncertainty}, as contextual entropy \citep{pimentelEtAl2023anticipation} and, more generally, as expectations of functions over sampled continuations \citep{giulianelli2024generalized}. Our model belongs to this family: the function is a grammar's score of how well a continuation completes the open goals, and the scores are turned into a next-event distribution. IAS, another member, measures the representational distance between a word and the alternative continuations a language model expected \citep{giulianelli2026ias}, and we evaluate it as a close alternative. Our weighting follows the logic of rescoring and fusion: an additional score changes how much probability the sequence model assigns to each continuation. This approach is established in controlled generation \citep{yangKlein2021fudge,zhao2024twisted,loulaEtAl2025smc}; here the score comes from goals inferred from the prefix, and the aim is to model human expectation rather than to generate text.
\section{Method}
\label{sec:method}

Grammar-Weighted Futures implements the hypothesis that expectations about how open goals will be completed constrain next-event probabilities. Given an observed prefix $h$, the chords or words heard so far, a fixed sequence model assigns a probability $q_E(c\mid h)$ to each possible next event $c$. Every probability model in this paper turns this sequence model's prediction into a hybrid prediction of the same form:
\begin{equation}
p_{\mathrm{hyb}}(c\mid h)=q_E(c\mid h)\bigl[(1-a_h)+a_h\Psi_h(c)\bigr],
\qquad \mathbb E_{q_E}[\Psi_h]=1.
\label{eq:correction}
\end{equation}
Here $\Psi_h(c)\geq0$ is an event-specific correction factor: events with $\Psi_h(c)>1$ become more probable and events with $\Psi_h(c)<1$ less probable. The gate $a_h\in[0,1]$, computed from the prefix, sets how strongly the correction is applied. Because $\Psi_h$ averages to one under $q_E$, $p_{\mathrm{hyb}}$ remains a probability distribution, and $\Psi_h\equiv1$ returns the sequence model's prediction unchanged.

Most comparisons in Section~\ref{sec:experiments} keep $q_E$ and $a_h$ fixed and change only how $\Psi_h$ is obtained, so they compare correction factors. H3, for example, compares the $\Psi_h$ obtained by scoring the sequence model's futures with one obtained from the grammar's own next-event probability (Section~\ref{sec:results-mechanism}). We describe our model for music, the primary test. It obtains $\Psi_h$ in three steps (Figure~\ref{fig:method}): it samples futures from the sequence model (\S\ref{sec:step-futures}), infers the grammar's open goals (\S\ref{sec:step-goals}), and scores how the futures complete them (\S\ref{sec:step-weights}). Sections~\ref{sec:step-gate} and~\ref{sec:language} define the gate and the language version.

\subsection{Step 1: sampling futures from a sequence model}
\label{sec:step-futures}

Human responses are recorded for \emph{events}: chords in music, and complete words or punctuation marks in language. The sequence model $q$ is autoregressive: it predicts one token at a time from the tokens before it, so it can also sample continuations $\omega$ of $h$, which we call futures. Futures are possibilities, not observed events. We write $\pi(\omega)$ for the first event of $\omega$, so that formally $q_E(c\mid h)=\Pr_q[\pi(\omega)=c\mid h]$.

In music, a Transformer, an LSTM and IDyOM, a variable-order Markov model that mixes corpus and current-piece statistics \citep{pearceWiggins2006expectation}, share a vocabulary of 739 chord symbols. Each symbol gives the chord's root (the note it is built on) relative to the song's annotated key, its quality (such as major, minor or seventh) and its inversion (which note is lowest), and each chord is one token, so $q_E$ is simply the model's next-chord distribution. The song's overall key is given to the models; brief moves to other keys within the song are left to the grammar. Apart from this key, everything the models use is computed from the prefix.

\subsection{Step 2: the grammar's open goals}
\label{sec:step-goals}

A probabilistic grammar is a set of rewrite rules with probabilities that generates sequences as trees. Both of our grammars are estimated from corpora, never from human responses (Appendix~\ref{app:grammar-pipeline}). The chord grammar, adapted from \citet{harasimEtAl2018parsing}, has 168 categories, the seven scale degrees (a chord's position within a key, written I--VII) in each of 24 major and minor keys. Its rules cover prolongation (repeating a chord's role), preparation (a chord that leads into another, such as V before I), tonicization (briefly treating another degree as a local home chord) and substitution (replacing a chord by another with the same role). Its rule probabilities are learned from the McGill Billboard chord corpus \citep{burgoyneEtAl2011billboard} without annotated trees.

Given $h$, an incremental parser finds the partial trees (analyses) that are consistent with the prefix and computes their probabilities \citep{jelinekLafferty1991computation,stolcke1995efficient}. In each analysis, the rules already used predict constituents (subtrees) that have not yet begun; their yields, the terminals they generate, lie in the future. These constituents are the prefix's \emph{open goals} (dotted in Figure~\ref{fig:method}a).

We track the first one or two open goals, the \emph{tracked goals}, and sum out later ones; two goals also capture what must follow the first (Appendix~\ref{app:frontier-example}). This gives posterior probabilities $p_1(g_1\mid h)$ for analyses with one open goal and $p_2(g_1,g_2\mid h)$ for ordered pairs of goals; $F$ denotes the first goal. Futures are scored up to a \emph{horizon} of $H$ terminals. The grammar also gives $d_h(\ell)$, its probability that completing the tracked goals takes $\ell$ terminals, renormalized over $1\leq\ell\leq H$ (Eq.~\ref{eq:duration}).

\subsection{Step 3: from scored futures to the correction}
\label{sec:step-weights}

At each length $\ell\leq H$, a sampled future receives a nonnegative score $G_\ell(\omega,h)$ for how its first $\ell$ terminals complete the tracked goals. In music, terminals are chords reduced to root and one of ten qualities, and $G_\ell$ is the probability that the first $\ell$ chords are exactly the yield of the tracked goals:
\begin{equation}
G^{\mathrm{music}}_\ell(\omega,h)=
\sum_{g_1}p_1(g_1\mid h)\,\beta_{g_1}(\omega_{1:\ell})
+\sum_{k=1}^{\ell-1}\sum_{g_1,g_2}p_2(g_1,g_2\mid h)\,
\beta_{g_1}(\omega_{1:k})\,\beta_{g_2}(\omega_{k+1:\ell}),
\label{eq:music-score}
\end{equation}
where $\beta_g(\cdot)$, the inside probability, is the probability that category $g$ generates exactly the given chords, and $k$ ranges over all ways of splitting the $\ell$ chords between the two goals. A future is thus scored by the qualities of its chords as well as by their roots.

Scores are then normalized within each length, so that they average to one over futures, and averaged over lengths with the weights $d_h(\ell)$. The result, $W_h(\omega)$, is the future's weight, and the correction factor for an event is the average weight of the futures that begin with it:
\begin{equation}
\Psi_h(c)=\mathbb E_q\bigl[W_h(\omega)\mid\pi(\omega)=c,h\bigr],\qquad
W_h(\omega)=\sum_{\ell=1}^{H}d_h(\ell)\,\frac{G_\ell(\omega,h)}{Z_\ell(h)},
\label{eq:conditional-value}
\end{equation}
with $Z_\ell(h)=\mathbb E_q[G_\ell(\omega,h)]$, so that $\mathbb E_q[W_h]=1$ as Eq.~\ref{eq:correction} requires. In words, an event becomes more probable when the futures that begin with it complete the goals well, and less probable when they leave the goals unmet. The resulting grammar-weighted next-event distribution is $p_{\mathrm{syn}}=q_E\Psi_h$. We estimate it from $N$ sampled futures by self-normalized importance sampling: for each length, $p_{\mathrm{syn}}(c\mid h)$ takes the score-weighted fraction of futures that begin with $c$, and mixes these fractions over lengths with weights $d_h(\ell)$ (Appendix~\ref{app:probability}, Eq.~\ref{eq:snis}; sample sizes in Appendix~\ref{app:surface-models}).

\subsection{Step 4: the gate}
\label{sec:step-gate}

The gate $a_h$ in Eq.~\ref{eq:correction} is computed from the grammar's analysis of the prefix. We selected the music gate form during LSTM development using the surprise ratings (Appendix~\ref{app:gate-development}). Equivalently, $p_{\mathrm{hyb}}=(1-a_h)\,q_E+a_h\,p_{\mathrm{syn}}$, a mixture of the original and grammar-weighted predictions. In music, the gate reflects how certain the prefix analysis is about the first goal $F$: $a_h=1-\mathbb H[p(F\mid h)]/\log_2 168$, where $\mathbb H$ is entropy in bits, so that $a_h$ is one when the analysis is certain of $F$ and zero when all 168 categories are equally likely. Both $q_E$ and $p_{\mathrm{hyb}}$ receive the same uniform smoothing, which spreads 3.4\% of the mass over all chords (Appendix~\ref{app:surface-models}).

\subsection{Language instantiation}
\label{sec:language}

Language uses the same construction with GPT-2 small \citep{radfordEtAl2019gpt2} as the sequence model and a probabilistic context-free grammar (PCFG) estimated from the parsed Penn Treebank (PTB) sample distributed with NLTK \citep{marcusEtAl1993ptb}, as in PCFG models of syntactic surprisal \citep{hale2001earley,levy2008expectation}. Three components change; Table~\ref{tab:implementations} in Appendix~\ref{app:grammar-pipeline} summarizes both instantiations.

\paragraph{Events.} GPT-2 predicts byte-pair-encoding (BPE) tokens and the grammar generates PTB terminals, neither of which always coincides with a word: one word may span several BPE tokens, and \emph{don't} corresponds to two PTB terminals. A lexer reads the next complete word or punctuation mark from each sampled continuation, and because a word can be spelled by several token sequences, $q_E$ sums over all of them \citep{caoRimell2021marginal,pimentelMeister2024word}; see Eq.~\ref{eq:word-event} in Appendix~\ref{app:probability}.

\paragraph{Scores.} Sentences rarely complete their goals within a few words, so a completion probability would give most futures zero weight. The language score instead credits progress: it is one plus the expected number of tracked goals that the future completes, over the analyses still consistent with the future, and zero if no analysis can continue it (Appendix~\ref{app:prefix-charts}). Future words enter only through their part-of-speech (POS) tags: a word-to-tag model trained on the same treebank gives each word a tag distribution, and the grammar's word probabilities are summed out because GPT-2 models word frequency. Normalizing these scores within each length and combining them as in Eq.~\ref{eq:conditional-value} gives the grammar-weighted next-word distribution, which the gate mixes with GPT-2's original prediction.

\paragraph{Gate.} Phrase-level goals constrain the next word mainly through its syntactic category, so the gate is the fraction of uncertainty about the next POS terminal $T$ that the first goal explains, $a_h=\mathbf 1\{m_h\geq4\}\,I(F;T\mid h)/\mathbb H[p(T\mid h)]$, where $I$ is mutual information and $m_h$ the number of PTB terminals in the prefix. The gate is zero when $T$ is certain or when the prefix has fewer than four terminals, because analyses of the first few words are unreliable (Appendix~\ref{app:prefix-charts}).

\section{Experiments}
\label{sec:experiments}

\subsection{Setup}
\label{sec:setup}

Musical surprise is the primary test of H1--H3, and language gives an exploratory test of H1. We measure how surprising each event is to a model by the information content (IC, $-\log_2 p(x\mid h)$, in bits) of the event $x$ that occurred. The \emph{Baseline} condition takes the IC from the sequence model's own prediction $q_E$, and \emph{Ours} takes it from the grammar-weighted prediction $p_{\mathrm{hyb}}$ of our model (Eq.~\ref{eq:correction}); each control changes one part of our model (Table~\ref{tab:conditions}). The IC then enters a separate response model, a regression of the human ratings on IC and other predictors. The grammar and sequence-model parameters are not fitted to human responses; we selected the music gate form by comparing alternatives on these ratings during LSTM development (Appendix~\ref{app:gate-development}).

\paragraph{How results are reported.} Each comparison asks which of two conditions better predicts human ratings that were held out from the response model. The measure is the expected log predictive density of the held-out ratings (ELPD, in nats), summed over all ratings, and a comparison reports the difference, $\Delta$ELPD; a positive value favors the condition named first. Two further numbers convey its size. Dividing $\Delta$ELPD by the number of ratings gives the average gain per rating: in music, a difference of $+20$ over the 2,018 ratings means that each held-out rating receives on average (geometric mean) about 1\% more predictive density. We also count the songs in which the first condition predicts better. Uncertainty is a 95\% interval from a paired bootstrap over the 30 songs (over sentences in language), and we call a difference \emph{reliable} when this interval excludes zero. Standard errors and the remaining intervals are given in the tables.

\paragraph{Music.} The surprise data of \citet{cheungEtAl2024expectancy} contain chord-by-chord surprise ratings of 30 chord progressions by 25 participants. We predict the mean rating of the musicians and of the nonmusicians at each of the 1,009 chords after the first, 2,018 ratings in all. The Transformer, LSTM and IDyOM are trained in ten song-level folds of the 739 McGill Billboard songs \citep{burgoyneEtAl2011billboard} used by \citet{cheungEtAl2024expectancy}, and each stimulus is scored by the models whose fold excludes its source song; the chord grammar is estimated once, from the 709 songs that are not stimulus sources (Appendix~\ref{app:grammar-conventions}). All three sequence models share the grammar, gate and smoothing, and futures are scored up to $H=10$ chords. For music predictors based on sampled futures, we use three independent sets of samples (streams), fit the response model separately for each stream and average the pointwise scores. The main analysis uses the response model released with the data and predicts each held-out rating from the other ratings in its song (conditional response-level leave-one-out, response-LOO). A second analysis holds out whole songs and predicts all their ratings from the other 29 songs (whole-song holdout; Appendix~\ref{app:response-models}). Results refer to the main analysis unless whole songs are said to be held out.

\paragraph{Language.} The resource of \citet{deVardaEtAl2024predictability} provides predictability ratings for 1,726 word events in 205 English sentences. We reverse the ratings, so that larger values align with greater IC, and predict them in ten folds that each hold out whole sentences. Ours combines released GPT-2 small \citep{radfordEtAl2019gpt2} with a PCFG and word-to-tag model estimated from the separate 3,914-sentence Penn Treebank sample distributed with NLTK \citep{marcusEtAl1993ptb}, scoring futures of up to $H=2$ PTB terminals. Developed on the same resource, this configuration is exploratory.

\paragraph{Transfer.} Secondary analyses test transfer to EEG recorded while listeners heard Bach chorales \citep{tengEtAl2024phrase} and to reading times, N400 \citep{frankEtAl2013reading,frankEtAl2015erp} and cloze (Appendix~\ref{app:language-followups}).

\begin{table}[t]
\centering\small
\caption{\textbf{Conditions compared in music.} Each condition forms its next-chord prediction by changing the sequence model's prediction $q_E$, and all share the same smoothing. Comparing a control with Ours answers the question in the last column.}
\label{tab:conditions}
\begin{tabular}{@{}lp{0.38\linewidth}p{0.36\linewidth}@{}}
\toprule
Condition & How the next-chord prediction is formed & Question answered by the comparison with Ours\\
\midrule
Baseline & The sequence model alone, $q_E$ & Does grammar weighting help? (H1)\\
Ours & $q_E$ reweighted by how well its sampled futures complete the open goals, with the gate & ---\\
Permuted & As Ours, but the scores are shuffled among the same sampled futures & Does it matter which future receives each score? (H2)\\
Entropy-matched & $q_E$ made as sharp or as flat as our model's prediction, without grammar scores & Is the gain only a sharper or flatter prediction? (H2)\\
\shortstack[l]{Grammar\\marginal} & $q_E$ multiplied by the grammar's own next-chord probability, with our model's gate & Do the sequence model's futures add to the grammar's next-chord prediction? (H3)\\
Static tilt & $q_E$ multiplied by a fixed preference over chord categories, with our model's gate & Is a context-independent preference for stable chords enough?\\
\bottomrule
\end{tabular}
\end{table}

\subsection{Grammar weighting and musical surprise (H1)}
\label{sec:results-h1}

\begin{figure}[t]
\centering
\includegraphics[width=\linewidth]{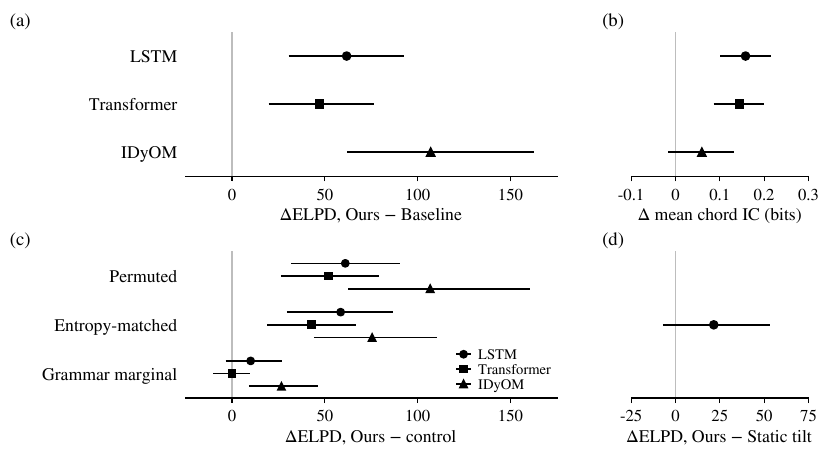}
\caption{\textbf{Musical surprise.} Points in (a), (c) and (d) show how much better Ours predicts held-out ratings than another condition ($\Delta$ELPD; positive values favor Ours). Whiskers are 95\% paired bootstrap intervals over the 30 songs, after averaging the three sampling streams. (a) Ours compared with Baseline, for each sequence model. (b) Change in the mean IC of the chords that occurred, Ours minus Baseline, in bits; positive values mean that Ours made these chords less probable. (c) Ours compared with each control (circle, LSTM; square, Transformer; triangle, IDyOM). (d) Ours compared with Static tilt for the LSTM, with whole songs held out.}
\label{fig:results}
\end{figure}

\emph{Does weighting futures by the grammar improve the prediction of listeners' surprise?} Yes, for all three sequence models (Figure~\ref{fig:results}a). Ours predicted the held-out ratings better than Baseline in 23 of the 30 songs for the LSTM, in 21 for the Transformer and in 27 for IDyOM. Per rating, Ours assigns on average 3.1\%, 2.4\% and 5.4\% more predictive density. All three gains are reliable, also when whole songs are held out (Appendix~\ref{app:mechanism}).

\emph{Does the gain come from predicting the chords that occurred better?} No (Figure~\ref{fig:results}b). For the two neural sequence models, Ours made the chords that actually occurred less probable, by 10.4\% and 9.5\% on average (geometric mean), and it did so in 23 and 25 of the 30 songs. For IDyOM, the change is not reliable. Ours thus predicts the ratings better without predicting the chords better.

\subsection{What produces the gain (H2 and H3)}
\label{sec:results-mechanism}

The next tests ask which part of our model produces the gain. Each control changes one part of it (Table~\ref{tab:conditions}) while retaining the response model and smoothing. We evaluate H2 and H3 for the LSTM, Transformer and IDyOM (Figure~\ref{fig:results}c; Appendix~\ref{app:mechanism}).

\paragraph{H2: does the assignment of grammar scores to futures matter?} Ours outperforms both controls. Any reweighting changes the sequence model's prediction, for instance by making it sharper, and such a change alone might improve the fit. Permuted reweights the same futures with the same set of grammar scores, but shuffled among them, so the scores no longer describe the futures they are attached to. Entropy-matched makes the sequence model's prediction as sharp or as flat as our model's without using the grammar. Neither control improves reliably on Baseline for the LSTM, and Ours is reliably better than both controls for all three sequence models (Figure~\ref{fig:results}c). The one qualification is IDyOM, for which sharpening alone recovers part of the gain (Appendix~\ref{app:mechanism}).

\paragraph{H3: do the sequence model's futures add to the grammar's own next-chord prediction?} The answer depends on the sequence model. Grammar marginal multiplies the sequence model's next-chord prediction by the grammar's own next-chord probability, which integrates possible completions of the open goals. It then applies the same gate and smoothing as Ours. The comparison asks whether scoring the sequence model's sampled continuations improves rating prediction beyond this correction.

For the LSTM, Grammar marginal alone recovers most of the gain ($+51.7$ of the $+61.7$ ELPD in Figure~\ref{fig:results}a). Ours is better than Grammar marginal in only 16 of the 30 songs, and the difference is not reliable ($+10.0$, $[-3.5,26.9]$). For the Transformer, the estimated difference is close to zero. For IDyOM, Ours predicts ratings reliably better than Grammar marginal (Figure~\ref{fig:results}c). Grammar marginal is not the same as Ours with futures of a single chord ($H=1$): that variant credits a future only when its first chord completes a goal by itself. For the two neural models, its estimated gain over Baseline is close to zero. With futures of two chords, Ours already reaches its full gain for the neural models (Appendix~\ref{app:mechanism}).

\paragraph{Statistical scoring controls.} Can simple corpus statistics reproduce the rating gains? We replace the goal-completion score of a continuation with a product of corpus bigram probabilities, one for each chord given its predecessor. We keep the sequence model, completion-length weights, gate and smoothing fixed. Ours predicts ratings better than this path control for the LSTM ($\Delta$ELPD $+56.4$, adjusted interval $[14.7,102.6]$) and Transformer ($+35.9$, $[12.2,60.0]$). We also test corrections that multiply the sequence model's next-chord probabilities by a chord preference, normalize, and apply the same gate and smoothing. Ours reliably outperforms a preference based on the last chord (Bigram tilt) for all three models. For the neural models, differences from fixed grammar preferences (Static tilt) and overall chord frequencies (Unigram tilt) remain uncertain. Appendix~\ref{app:postreview-music} reports the full comparisons, a test of learned grammar probabilities, and a length-matched comparison with Grammar marginal.

\subsection{Language (exploratory)}
\label{sec:results-language}

\emph{Does grammar weighting also help in language?} Over GPT-2 alone, yes. Ours predicted the reversed predictability ratings reliably better than Baseline, and did so in 114 of the 205 sentences (Table~\ref{tab:language}). Four simpler syntax measures, each added as a predictor beside Baseline, do not account for this gain (Appendix~\ref{app:language-results}). The gain is largest for futures of $H=2$ PTB terminals; since some rated words span several terminals, this peak does not show that readers anticipate several words ahead.

\emph{Does it add to the closest future-based alternative?} Not detectably. IAS \citep{giulianelli2026ias} measures the representational distance between a word and the alternative continuations a language model expected. Its released score improves prediction more than Ours does, and adding Ours on top of it gives a positive but uncertain further gain (Table~\ref{tab:language}; nested-selection variants in Appendix~\ref{app:language-results}). The result thus supports grammar weighting over GPT-2 alone, not over IAS.

\begin{table}[t]
\centering
\small
\caption{Exploratory language results: how much better the first condition predicts held-out ratings ($\Delta$ELPD with sentence-cluster SE, and 95\% paired bootstrap interval over the 205 sentences). Each of the ten folds holds out whole sentences.}
\label{tab:language}
\begin{tabular}{lrr}
\toprule
Contrast & $\Delta$ELPD $\pm$ SE & 95\% interval\\
\midrule
Ours $-$ Baseline & $+5.080\pm2.032$ & $[1.188,9.088]$\\
(Baseline + released IAS) $-$ Baseline & $+9.574\pm4.509$ & $[0.629,18.328]$\\
(Baseline + released IAS + Ours) $-$ (Baseline + released IAS) &$+3.671\pm4.910$ & $[-5.981,13.144]$\\
\bottomrule
\end{tabular}
\end{table}

\subsection{Transfer beyond explicit ratings}
\label{sec:results-transfer}

\emph{Does the gain carry over to other measures of expectation?} Not reliably. In the EEG data, modeled with two LSTM sequence models, chord IC from Baseline or Ours did not improve held-out prediction beyond acoustic features for either model, and Ours did not differ from Baseline. Reading times and cloze show no reliable gain, and the N400 gain depends on analysis settings (Appendix~\ref{app:language-followups}).

\section{Discussion}
\label{sec:discussion}

Grammar weighting improved the prediction of listeners' surprise ratings over Baseline for all three sequence models, including on held-out songs.

The gain did not come from predicting the chords better. For the neural models, our model made the chords that occurred about 10\% less probable, yet it predicted the ratings better. Similar dissociations are known in language, where models that predict words better need not fit reading behavior better \citep{kuribayashiEtAl2021lower,ohSchuler2023poorer}, and language-model surprisal underestimates syntactic processing difficulty \citep{vanSchijndelLinzen2021single,arehalliEtAl2022syntactic,huangEtAl2024benchmark}. Unlike those studies, we held the sequence model fixed, so the dissociation arises from reweighting its predictions.

Ours outperformed shuffled grammar scores and entropy matching (H2). Goal-completion scores also gave better rating predictions than corpus bigram scores for the neural models when both were used to score entire continuations. A fixed preference for chords that are stable in a key \citep{bharuchaKrumhansl1983stability,lerdahl2001tonal} recovered more than half of the LSTM's gain on held-out songs (Figure~\ref{fig:results}d). The static and unigram next-chord controls retain the grammar-derived gate, and their uncertain differences from Ours leave open how much of the neural-model gain comes from scoring the particular goals inferred in each progression (Appendix~\ref{app:postreview-simple}).

For the LSTM and Transformer, Grammar marginal recovered most of the improvement in rating prediction over Baseline. It retains the sequence model's next-chord probabilities and corrects them with the grammar's predictive probabilities, computed by a prefix parser that integrates possible completions of the open goals \citep{hale2001earley,levy2008expectation,stolcke1995efficient,roark2001probabilistic}. Ours instead derives the correction by scoring the sequence model's sampled continuations, with the same gate and smoothing. The sequence model's sampled continuations yielded a reliable gain over Grammar marginal for IDyOM. One possibility is that Grammar marginal already captures the distinctions between chord functions that matter most for chord-by-chord surprise. Scoring continuations drawn from the sequence model can shift these next-chord preferences without bringing them closer to listeners' ratings.

These findings have clear bounds. The main evidence comes from one study of 30 popular-song chord progressions, rated chord by chord, with each song's key given to the models. The whole-song static-prior comparison uses the LSTM. The language result is exploratory and limited to one GPT-2/PCFG configuration on 205 rated sentences, with a grammar that sees future words only through part-of-speech tags, and the gain did not transfer reliably to EEG or reading times (Section~\ref{sec:results-transfer}). Our claim therefore concerns explicit judgments of surprise. Further language comparisons could test recurrent neural network grammars \citep{haleEtAl2018finding} and incremental predictive parsers \citep{demberg2013pltag}, and use Future Lens \citep{palEtAl2023futurelens} to examine whether the sequence model itself represents open goals.

Grammar-Weighted Futures makes structural anticipation computationally testable by evaluating a learned sequence model's possible futures against explicit open goals. For the neural models, this weighting improves the prediction of listeners' surprise while reducing next-chord likelihood, so the two evaluation criteria favor different output distributions from the same trained network.

\section{Conclusion}
\label{sec:conclusion}

This paper addressed a long-standing question: does expectation in language and music rest only on the probability of the next event, or also on structure that the context has left open? Surprisal theory and statistical learning have made next-event probability the common currency of expectation \citep{hale2001earley,levy2008expectation,pearceWiggins2012auditory}, whereas structural theories hold that readers and listeners also track open goals \citep{gibson1998locality,rohrmeier2013expectancy}. Direct comparisons between sequential and hierarchical models have been hard to interpret, because the models differ in more than their structure \citep{frankBod2011insensitivity,fossumLevy2012sequential}. \emph{Grammar-Weighted Futures} instead adds a grammar's open goals to a trained sequence model without retraining it, so that the predictive contribution of grammar weighting can be measured while the sequence model stays fixed. For three such models, grammar weighting made listeners' surprise ratings more predictable. The relation between learned statistics and structural expectation thus becomes a question that can be tested with other grammars, models and data.

\section*{AI Usage Disclosure} We used Claude Code and OpenAI Codex to assist with conceptual and hypothesis refinement, mathematical exposition and checks, experimental design and parameter suggestions, method implementation, debugging, data processing, and data analysis. These tools also supported literature discovery, figure preparation, manuscript editing, and pre-submission critique. The authors determined the research questions, final experimental design, and scientific claims. They personally reviewed the code and tests, checked mathematical derivations and numerical results, verified cited references against original sources, and reviewed the interpretations and final manuscript against the evidence. AI-generated material was not used as human behavioral or neural observations. The authors take full responsibility for the content of this work, including its text, claims, analyses, and artifacts.

\section*{Ethics Statement} This study reanalyzes the cited behavioral, EEG and corpus resources; it collects no new participant data. Human-subject recruitment and consent are described in the original studies. We do not claim a new ethics approval or permission to redistribute the original recordings and participant data.

\section*{Reproducibility Statement} Section~\ref{sec:method} and Appendices~\ref{app:probability}--\ref{app:response-models} describe the estimators, grammar construction, sequence models, data splits, and evaluation procedures. Section~\ref{sec:setup} identifies the source datasets. Appendix~\ref{app:postreview-music} gives the methods and full comparison tables for the additional music controls. Appendix~\ref{app:reproducibility} reports numerical checks for the music analyses.

\bibliography{references}
\bibliographystyle{iclr2027_conference}

\clearpage \appendix
\section{Probability construction and estimator checks}
\label{app:appendix-start}
\label{app:probability}
\label{app:estimator-validation}

\paragraph{Word-event probability.}
Each language prefix ends at a word boundary, so the next word must begin with whitespace. With $q_0$ the GPT-2 sampling distribution (end-of-sequence suppressed) and $B$ the event that the continuation begins with whitespace,
\begin{equation}
q_E(c\mid h)=
\frac{\mathbb E_{q_0}[\mathbf 1_B\mathbf 1\{\pi(\omega)=c\}]}{\Pr_{q_0}(B\mid h)},
\label{eq:word-event}
\end{equation}
and the sequence model's distribution is correspondingly $q=q_0(\cdot\mid B,h)$. This sums over all BPE spellings of a word rather than reading one softmax entry \citep{caoRimell2021marginal,pimentelMeister2024word}. Rare rated words are seldom sampled, so an additional sample of futures that begin with the rated word is drawn and combined with the ordinary sample by stratified estimation.

\paragraph{Built-in null.}
If every $G_\ell=1$, then $\Psi_h\equiv1$ for any $H$ and $d_h$, because a normalized autoregressive sampling law returns its first-token marginal:
\[
\Pr_q(\omega_1=c\mid h)=q(c\mid h)
\sum_{\omega_{2:H}}\prod_{j=2}^Hq(\omega_j\mid h,\omega_{<j})=q(c\mid h).
\]
With finitely many futures this holds only up to sampling error, so in music, where $q_E$ is the exact softmax, the two can differ for rare events. Independent first-token counts have binomial variance, and $\mathbb E\|\widehat q_N-q\|_2^2=(1-\sum_cq(c)^2)/N$. The reference must match the actual sampler: our GPT-2 cache suppresses EOS. For decoded words the identity instead concerns the pushforward under $\pi$: an initial token spelling \texttt{\_mark} can extend to \emph{marker}, so its softmax probability is not the probability of the closed word \emph{mark} \citep{pimentelMeister2024word}. Both arms use the same target-blind whitespace-boundary event $B$ and word projection.

\paragraph{From scores to next-event probabilities.}
Equivalently to Eq.~\ref{eq:conditional-value}, $p_{\mathrm{syn}}=q_E\Psi_h$ mixes over lengths the next-event marginals of the grammar-tilted distributions $r_\ell(\omega\mid h)\propto q(\omega\mid h)G_\ell(\omega,h)$:
\begin{equation}
p_{\mathrm{syn}}(c\mid h)=q_E(c\mid h)\Psi_h(c)
=\sum_{\ell=1}^{H}d_h(\ell)\,
\frac{\mathbb E_q[\mathbf 1\{\pi(\omega)=c\}\,G_\ell(\omega,h)]}
     {\mathbb E_q[G_\ell(\omega,h)]}.
\label{eq:first-marginal}
\end{equation}
Lengths are weighted by $d_h(\ell)$ rather than by $d_h(\ell)Z_\ell(h)$, as joint normalization over length and future would do. This modeling choice keeps the expected completion length separate from the scale of the music scores, which shrink as more rule probabilities are multiplied.

\paragraph{SNIS estimator.}
With $N$ independent futures $\omega^{(i)}\sim q$, each length term of Eq.~\ref{eq:first-marginal} is estimated as
\begin{equation}
\widehat p_{\mathrm{syn},\ell}(c\mid h)
=\frac{\sum_{i=1}^N G_\ell(\omega^{(i)},h)\,\mathbf 1\{\pi(\omega^{(i)})=c\}}
       {\sum_{i=1}^N G_\ell(\omega^{(i)},h)}.
\label{eq:snis}
\end{equation}
Because the proposal is $q$ itself, the importance weights are the scores; weighting again by $q$ would count frequent futures twice. In language, futures that do not begin at a word boundary receive zero weight. The estimator has finite-sample ratio bias but is consistent under the usual support and moment conditions \citep{owen2013montecarlo}. Every length had positive estimated support in the reported analyses.

\paragraph{Monte Carlo precision.}
In 100 paired resamplings of the language caches, with the complete behavioral procedure refitted each time, the Ours~$-$~Baseline contrast has SD $.038$ ELPD at $H=10$ and $.058$ at $H=2$, against sentence SEs of $1.392$ and $2.032$.

\section{Grammar and surface implementations}
\label{app:grammar-pipeline}
\label{app:training}

Table~\ref{tab:implementations} summarizes how the construction of Section~\ref{sec:method} is instantiated in music and language.

\begin{table}[ht]
\centering\small
\caption{One probability operator, two structural instantiations. Neither grammar is fitted to human ratings.}
\label{tab:implementations}
\begin{tabular}{p{.17\linewidth}p{.33\linewidth}p{.41\linewidth}}
\toprule
Component & Music & Language \\
\midrule
Surface event & One chord symbol & Complete word/punctuation event \\
Grammar terminal & Chord root and quality & Part-of-speech tag \\
Open goal & Scale degree in a local key & Phrase category (e.g., VP) \\
Score $G_\ell$ & Probability of completing the goals & Progress toward completing the goals \\
Gate & Certainty about the first goal & How well the first goal predicts the next tag \\
\bottomrule
\end{tabular}
\end{table}

\subsection{Chord grammar: rules and estimation}
\label{app:grammar-conventions}

A category $X_{d,\kappa}$ denotes degree $d$ in local key $\kappa=(r,m)$. Degree labels are indices, not chord qualities. Roots use
\begin{equation}
\operatorname{root}(d,\kappa)=(r+\delta_m(d))\bmod12,\quad
\delta_{\rm maj}=(0,2,4,5,7,9,11),\quad
\delta_{\rm min}=(0,2,3,5,7,8,10).
\label{eq:grammar-root}
\end{equation}
Let $\mu(d,\kappa)=(\operatorname{root}(d,\kappa),\eta_m(d))$. The fixed target modes are \[
\begin{split}
\eta_{\rm maj}&=(\mathrm{maj,min,min,maj,maj,min,min}),\\
\eta_{\rm min}&=(\mathrm{min,min,maj,min,maj,maj,maj}).
\end{split}
\] The minor fallbacks for diminished degrees are modeling conventions, not established modulations; they are broader than the restriction in \citet{rohrmeier2011generative}. For $d\ne\mathrm I$, tonicization and dominant preparation are
\begin{equation}
X_{d,\kappa}\to X_{\mathrm I,\mu(d,\kappa)},\qquad
X_{d,\kappa}\to X_{\mathrm V,\mu(d,\kappa)}X_{d,\kappa}.
\label{eq:grammar-tonicization}
\end{equation}
Remaining schemas, adapted from \citet{harasimEtAl2018parsing}, are
\begin{align*}
X_{d,\kappa}&\to X_{d,\kappa}X_{d,\kappa},&
X_{d,\kappa}&\to X_{d+4,\kappa}X_{d,\kappa}\quad(d\ne\mathrm{IV}),\\
X_{\mathrm I,\kappa}&\to X_{\mathrm{IV},\kappa}X_{\mathrm I,\kappa},&
X_{\mathrm I,(r,m)}&\to X_{\mathrm I,(r,\bar m)},\\
X_{\mathrm I,(r,\rm maj)}&\to X_{\mathrm{VI},(r,\rm maj)},&
X_{\mathrm I,(r,\rm min)}&\to X_{\mathrm{III},(r,\rm min)},\\
X_{\mathrm{II},\kappa}&\to X_{\mathrm{IV},\kappa},&
X_{\mathrm V,\kappa}&\to X_{\mathrm{VII},\kappa},\\
X_{\mathrm V,(r,m)}&\to X_{\mathrm V,(r+s\bmod12,m)},&&s\in\{3,6,9\}.
\end{align*}
Degree addition is modulo seven with I indexed by zero; $\bar m$ is the opposite mode. Each degree/mode distribution jointly covers structural rules and ten unrestricted chord-form emissions: major, major seventh, major sixth, dominant seventh, minor, minor seventh, half-diminished seventh, diminished seventh, augmented and suspended. Probabilities are tied across tonic roots. Unlike theory-constrained termination, no degree-specific form exclusions apply. Natural-minor VII emits on the subtonic; uppercase V does not force a major chord. The grammar omits inversions, voice leading, meter and cadence strength.

Training expresses each fragment relative to the annotated tonic of its region. The root category of a training fragment is drawn from a fixed uniform prior over the fourteen degree--mode categories of that tonic (seven degrees in major and in minor), so fragments need not begin on the tonic; this prior is not learned. Prediction uses a fixed equal start mixture over $X_{\mathrm I,(0,\rm maj)}$ and $X_{\mathrm I,(0,\rm min)}$; the tonic itself is not inferred over all keys. Evaluation inherits source-aligned tonal coordinates, constant within each analyzed excerpt. Latent local keys inside parses are distinct from this supplied reference and do not establish a perceptual modulation. Adjacent duplicates collapse in training, and fragments are not truncated. Evaluation preserves response positions, including 34 adjacent pairs merged by grammar projection.

Trees are latent. A single grammar is estimated from the 709 corpus songs that are not sources of any Cheung stimulus: 6,983 fragments with 68,919 chords after collapsing adjacent duplicates. We use a fully factorized Dirichlet variational posterior $Q$ over the tied rule distributions \citep{kuriharaSato2006variational}. Each fragment, one occurrence of an annotated song section within one tonic region, is a complete, finite observation: its latent derivation, rooted in the fixed fragment prior, must generate exactly the observed chords. One deterministic full-batch coordinate update first recomputes the optimal local tree distribution for every training fragment from the current expected log rule weights, using exact inside expected counts, and then sets
\begin{equation}
Q(z_i)\ \propto\ \exp\!\left\{\mathbb E_{Q(\theta)}
  [\log p_\theta(x_i,z_i)]\right\},\qquad
\nu_r\leftarrow\alpha+\sum_{i\in\mathcal D}
  \mathbb E_{Q(z_i)}[n_r(z_i)].
\label{eq:music-cavi}
\end{equation}
Here $\alpha=1$ and $\mathcal D$ contains every training fragment. Length-dependent microbatches only bound memory: their expected counts are accumulated at one fixed global state before the unit-step update, so they do not make the algorithm stochastic. Updates stop once the gain in the evidence lower bound falls below $10^{-5}$ nats per training chord in two consecutive updates, which happened after 128 updates; no behavioral or neural outcome enters training or stopping. Prediction uses normalized posterior-mean rule probabilities. The implementation checks support before training and count conservation and monotone bound improvement during training. After training it checks that the grammar is subcritical, with offspring spectral radius $0.960$, so that derivations terminate with probability one, and that the 350 held-out fragments of the 30 source songs have finite scores. No gold trees, test songs or human behavioral or neural responses enter estimation.

\subsection{Prefix charts and language adapter}
\label{app:prefix-charts}

Both domains use unit closure $R_U=(I-P_U)^{-1}$ and left-corner closure $R_L=(I-P_L)^{-1}$ on original categories. Music uses $P_L=R_UP_{\rm left}$; language includes the first child of every production. Their spectral radii are below one for the reported grammars. Language's unit and left-corner radii are $.015$ and $.30$, and its termination is certified by extinction iteration. No category-changing grammar transform is applied. A span chart combines boundary-completion and split terms, closed under $R_L$. Normalizing by continuing-analysis mass treats the prefix as incomplete. A second chart and top-down adjoint yield one-goal and ordered two-goal masses, summing out deeper goals. The first open goal $F$ has marginal $p(F=x\mid h)=p_1(x\mid h)+\sum_y p_2(x,y\mid h)$; this marginal and the total mass are checked.

Language uses the relative-frequency PCFG of 3,914 NLTK PTB trees: 72 nonterminals, 114 unary, 511 binary and 3,130 longer productions, without binarization, and unknown-word pseudo-count $.05$. Prefixes use lexical emissions. Future projection preserves structural rules but sums out lexical emissions; a word-form adapter provides independent soft POS evidence $p(t\mid w)$ with rare-word backoff. This posterior evidence reintroduces a tag-frequency prior and is an approximation, not a lexical likelihood or unsupervised discovery of syntax. After each future terminal, the chart divides the mass of the analyses still consistent with the future, $m_s(\omega,h,\ell)$, among four progress states $s$: the first goal is still open, the first of two goals is complete, the only goal is complete, or both are complete. The score is $G^{\mathrm{lang}}_\ell=\sum_su_sm_s/\sum_sm_s$ with utilities $u=(1,2,2,3)$, one plus the expected number of goals completed, or zero when compatible mass vanishes. Averaged over events and over lengths weighted by $d_h$, about three quarters of the compatible mass still has a tracked goal open, which is why language scores progress rather than completion. The score measures progress rather than probability, so a rare compatible path scores the same as a common one; a future also scores zero once it runs past both completed goals. Music instead uses exact-yield inside probabilities (Eq.~\ref{eq:music-score}). Replacing yield probabilities by total length masses gives the completion-length distribution
\begin{equation}
d_h(\ell)\;\propto\;\sum_{g_1}p_1(g_1\mid h)\,\Pr\nolimits_{\mathcal G}\bigl(|\mathrm{yield}(g_1)|=\ell\bigr)
+\sum_{g_1,g_2}p_2(g_1,g_2\mid h)\,\Pr\nolimits_{\mathcal G}\bigl(|\mathrm{yield}(g_1g_2)|=\ell\bigr),
\label{eq:duration}
\end{equation}
the probability under the grammar $\mathcal G$ that completing the tracked goals takes $\ell$ terminals, truncated at $H$ and renormalized over $1\leq\ell\leq H$. Before this renormalization, the mass within $H=10$ averages 74.5\% (median 74.1\%) in music and 74.4\% in language. Gate statistics use the first-goal posterior; language additionally uses $P(T\mid F=x)=[R_LE]_{x,t}$, with $E$ the preterminal-emission matrix. The language gate's threshold of four prefix terminals was chosen during development by the log loss of the actual next words on training sentences, never by ratings.

\paragraph{Worked open goals.}
\label{app:frontier-example}
Take $S\to\mathrm{SUBJ}\ \mathrm{VP}\ \mathrm{STOP}$, $\mathrm{SUBJ}\to\text{Alice}$, $\mathrm{ADV}\to\text{quickly}$, $\mathrm{STOP}\to\text{.}$, and equiprobable $\mathrm{VP}\to\text{runs}\mid\mathrm{ADV}\ \mathrm{VP}$. After \emph{Alice}, goals are $(\mathrm{VP},\mathrm{STOP})$. \emph{runs .} completes VP and then STOP (utilities $2,3$); \emph{quickly runs} leaves VP open and then completes it ($1,2$). With equal proposal mass and $H=2$, the truncated duration mass is all at two, so structural weighting changes $\Pr(\text{runs})$ from $1/2$ to $3/5$. Exact completion instead scores these paths $1/2$ and zero. This toy is not a PTB parse: its one-terminal prefix has zero maturity gate, so $p_{\mathrm{hyb}}$ equals $q_E$.

\subsection{Sequence models and execution}

\label{app:surface-models}

Music uses a four-layer, four-head pre-normalization Transformer (embedding 256, feed-forward 1,024) and a three-layer LSTM (embedding 256, hidden 384), context up to 128 tokens, and next-chord cross-entropy. Ten duplicate-grouped folds partition 739 Billboard songs; the grammar is instead estimated once from the 709 songs that are not sources of any stimulus (Appendix~\ref{app:grammar-conventions}). Training duration is chosen within the training partition before reinitialization and full-partition fitting. Each test excerpt uses the fold excluding its source song. IDyOM combines long-/short-term models by entropy-weighted geometric pooling; hypothetical events never update counts. $q_E$ and $p_{\mathrm{hyb}}$ receive the same uniform smoothing mass, $36/1060$ spread over the 739 symbols, which keeps surprisal finite and was not tuned on human data. Language uses frozen pretrained GPT-2 and cached proposals.

Both domains track two goals. Music uses $H=10$ chords, $N=16{,}384$ and three streams (seeds 11, 22, 33); language uses $H=2$ PTB terminals, $N=4{,}096$ ordinary and $1{,}024$ target-conditioned futures, with a sweep over $H=1,\dots,10$ as sensitivity. PTB terminals are not words or BPE tokens: 240 target words map to multiple terminals. Language starts with 80 generated BPE tokens (after the forced prefix for target draws), allowing up to 4,096 additional tokens. Censored paths retain reached-length scores and zero at unreached lengths, without changing denominators. One ordinary path remains censored; every positive-duration event/length cell has recorded support. Per-prefix grammar scoring takes about $.5$ seconds for music's three streams and $2.6$ seconds for language on one GPU.

\section{Response models and predictive scoring}
\label{app:response-models}

\subsection{Music: represented-song and unseen-song targets}

\paragraph{Gate development.}
During LSTM development, we compared eleven gate variants on the same surprise ratings used in the music evaluation. These included constant mixture weights and gates that vary with the prefix. We selected the gate based on certainty about the first open goal, which had the highest leave-one-out rating prediction score in this comparison. The analysis in Appendix~\ref{app:mechanism} compares this gate with a constant set to its mean across prefixes.
\label{app:gate-development}

Surprise data from 25 participants yield 2,018 musician-group means at 1,009 non-initial positions across 30 songs \citep{cheungEtAl2024expectancy}. We follow the response model in the authors' released analysis code, replacing only its focal IC predictor. The Gaussian identity-link population design is $\mathrm{VAL}+\mathrm{ARO}+\mathrm{musician}*\mathrm{IC}$, with VAL and ARO the valence and arousal ratings; all six coefficients including intercept vary jointly by song with covariance $\Sigma=\operatorname{diag}(\tau)\Omega\operatorname{diag}(\tau)$. These song effects are partially pooled rather than estimated independently. Continuous variables use full-panel mean/SD standardization; musicianship is binary. Within 60 song-by-musicianship series, residuals satisfy $r_i-\rho r_{\mathrm{prev}(i)}\sim\mathcal N(0,\sigma^2)$, with $r_i\sim\mathcal N(0,\sigma^2)$ at series start (nonstationary initialization). Thus $\rho$ accounts for serial dependence between adjacent chord ratings, while shared song effects link the two musicianship groups.

The response model for surprise uses $\mathcal N(0,1)$ fixed-effect priors, half-Cauchy$(0,1)$ scale priors, LKJ$(2)$ correlations and uniform $\rho\in(-1,1)$. Four chains retain 2,500 of 5,000 iterations each, with adaptation target $.99$ and maximum tree depth 10.

The released analysis applies \texttt{brms} LOO to the same fitted model. We retain that response model but, because its observations are correlated, make our predictive target explicit and evaluate the conditional density of a missing response given the remaining responses in its song. For this conditional response-LOO, integrate song effects. Let $A_g$ have diagonal one and subdiagonal $-\rho$ within each series; $\mu_g=X_g\beta$ and \[
\begin{gathered}
V_g=Z_g\Sigma Z_g^\top+\sigma^2A_g^{-1}A_g^{-\top},\qquad Q_g=V_g^{-1},\\
f_i(\psi)=\mathcal N\!\left(y_i;\ y_i-
\frac{[Q_g(y_g-\mu_g)]_i}{[Q_g]_{ii}},\ [Q_g]_{ii}^{-1}\right).
\end{gathered}
\] The last argument is variance; the mean cancels $y_i$. PSIS uses inverse-density ratios $1/f_i(\psi)$ from full-posterior global draws $\psi=(\beta,\Sigma,\sigma,\rho)$ \citep{buerknerEtAl2021nonfactorized,liEtAl2016integrated}. This predicts an omitted response given the other responses in its song, not an unseen song. Contributions with Pareto $k>0.7$ or pointwise MCSE $>0.03$ receive exact held-out refits that integrate the missing response while preserving AR indices; priors and preprocessing stay fixed. The Pareto-$k$, MCSE and refit diagnostics for the main music fits are summarized in Appendix~\ref{app:reproducibility}.

Paired event differences $\delta_i$ are averaged across streams first. Song-cluster SE is $\sqrt{30\operatorname{Var}_g(\sum_{i\in g}\delta_i)}$, using sample variance; bootstrap intervals resample songs. This is stimulus dispersion, not computation or selection uncertainty. The separate whole-song holdout fits 29 songs by Gaussian maximum likelihood and jointly predicts the thirtieth, with training-only scaling, the same fixed effects, and random intercept plus each arm's own IC slope. Every whole-song fit converged. Its ELPD is a different target.

\subsection{Language and secondary outcomes}

The reading times, N400 amplitudes and released PSG surprisal values originate from \citet{frankEtAl2013reading,frankEtAl2015erp} and are distributed aligned with the ratings by \citet{deVardaEtAl2024predictability}. Language uses deterministic ten-fold sentence-held-out validation with training-only standardization and frequency imputation. $X_s$ contains intercept, position, SUBTLEX log frequency, missing-frequency indicator, word length and focal IC; $Z_s$ contains intercept and IC. The training covariance is $V_s=Z_s\Sigma Z_s^\top+\sigma^2\Gamma_s(\rho)$, with unrestricted $2\times2$ $\Sigma$ and $[\Gamma_s]_{ij}=\rho^{|\mathrm{position}_i-\mathrm{position}_j|}$. Profiled maximum likelihood uses two starts and $\rho=.95\tanh\theta$. With $C_\beta=(\sum_{\rm train\ s}X_s^\top V_s^{-1}X_s)^{-1}$, the held-out density is \[
y_s\sim\mathcal N\!\left(X_s\widehat\beta,
Z_s\widehat\Sigma Z_s^\top+\widehat\sigma^2\Gamma_s(\widehat\rho)+X_sC_\beta X_s^\top\right).
\] This integrates unseen-sentence effects and approximate fixed-effect uncertainty, conditional on fitted covariance and predictors. Natural-log joint densities are not independent word scores or exact sentence LOO. Same-fold response Jacobians cancel in contrasts. SEs and 10,000-draw percentile intervals use 205 sentences; cache-resampling precision is evaluated separately (Appendix~\ref{app:probability}). Duration selection uses five sentence-grouped inner folds over 27 rules. Neither removes earlier dataset-level development.

The music EEG data \citep{tengEtAl2024phrase} contain 62-channel recordings from 29 listeners who heard Bach chorales at three tempi. We use the original-order trials of nine chorales, excluding one whose audio does not match its reference score, which leaves 772 of 783 participant-trials after omitting, in every arm, trials with at most 500 artifact-free samples. Ocular artifacts are removed by ICA learned from non-music calibration periods. The 576 chord events come from expert Roman-numeral analyses \citep{gothamEtAl2023whenInRome} aligned to the rendered beats at 100~Hz; labels follow the audio where it departs locally from the score, and a sensitivity variant keeps the score's labels. Two factorized LSTM sequence models predict them: one trained in ten folds on Billboard and one adapted to 339 expert-annotated Bach chorales that exclude the stimuli and their tune families. Both use the chord grammar, $H=10$, gate and smoothing of the main analysis, with probabilities averaged over three streams of 16,384 futures before IC and predictive entropy are computed from the root-and-quality marginal (primary) and from the root and full-chord marginals. The TRF regresses each channel on 0--600~ms lags of the cochlear envelope and of beat and chord-onset impulses (together, the acoustic features), plus IC and entropy impulses at chord onsets. Nine outer folds hold out all tempi of one chorale; four training-only validation chorales tune separate acoustic and chord penalties, including omission of the chord block. Held-out correlation $r$ is averaged over channels and trials with chorales weighted equally, and participant-level paired differences receive 10,000-draw bootstrap intervals and Holm correction within each family of contrasts. Additional language outcomes and their limits appear below.

\section{Controls, sensitivity, and transfer}

Numerical checks are summarized in Appendix~\ref{app:reproducibility}.

\subsection{Music: alternatives to dynamic completion}
\label{app:mechanism}
\label{app:blind-music-controls}

Conditional response-LOO controls use all 1,009 events and three proposal streams (Table~\ref{tab:music-controls}). Grammar marginal, Permuted and Entropy-matched are evaluated for all three sequence models. Permuted preserves each prefix--length set of scores, Grammar marginal retains the same grammar and gate but bypasses sampled futures, and Constant gate replaces the event-specific gate by its mean across prefixes, $0.620$. The event-specific gate adds little over its mean (Ours~$-$~Constant gate, $+1.79\pm3.14$).

Across sequence models, Ours exceeds both H2 controls, while Ours~$-$~Grammar marginal ranges from about zero for the Transformer to a resolved gain for IDyOM. For IDyOM, Entropy-matched alone improves on Baseline ($+31.56\pm13.89$, interval $[6.74,59.82]$), so part of its gain reflects the sharpness of its prediction; Ours still exceeds Entropy-matched by $+75.45\pm17.29$. For the Transformer, Permuted is slightly worse than Baseline ($-4.59\pm2.14$, $[-9.02,-0.73]$).

\paragraph{Horizon.} Varying $H$ from 1 to 10 with the same grammar, gate, streams and response-LOO shows a step from $H=1$ to $H=2$ and a plateau thereafter. Ours~$-$~Baseline at $H=1$ is $+2.54\pm31.05$, $-1.21\pm28.73$ and $+44.16\pm34.29$ for the LSTM, Transformer and IDyOM, and at $H=2$ it is $+62.36\pm15.98$, $+48.35\pm16.71$ and $+98.29\pm25.45$. No horizon exceeds $H=10$ by more than $+2.93\pm4.12$ (Transformer, $H=3$; interval $[-3.91,11.86]$). Grammar marginal exceeds Ours with $H=1$ by $+49.17\pm32.78$, $+48.36\pm26.15$ and $+36.25\pm26.30$. The next-chord distribution converges in the same way for all three sequence models: its mean total variation from the $H=10$ distribution is $0.335$, $0.325$ and $0.332$ at $H=1$, $0.067$, $0.069$ and $0.068$ at $H=2$, and at most $0.034$ from $H=3$ on, whereas Ours with $H=10$ differs from Baseline by $0.194$, $0.207$ and $0.221$.

\begin{table}[ht]
\centering\small
\caption{Controls under conditional response-LOO ($\Delta$ELPD $\pm$ song-cluster SE, with 95\% paired song-bootstrap interval). Positive values favor the first condition.}
\label{tab:music-controls}
\begin{tabular}{lll}
\toprule
Condition & Condition $-$ Baseline & Ours $-$ condition\\
\midrule
\multicolumn{3}{l}{\emph{LSTM}}\\
Grammar marginal & $+51.71\pm14.19$ $[24.46,79.15]$ & $+10.03\pm7.92$ $[-3.48,26.88]$\\
Permuted & $+0.73\pm2.33$ $[-3.34,5.69]$ & $+61.02\pm15.20$ $[31.93,90.30]$\\
Entropy-matched & $+3.23\pm6.16$ $[-8.00,15.52]$ & $+58.52\pm14.77$ $[29.60,86.98]$\\
Constant gate & $+59.95\pm15.79$ $[29.30,89.98]$ & $+1.79\pm3.14$ $[-4.71,7.51]$\\
\midrule
\multicolumn{3}{l}{\emph{Transformer}}\\
Grammar marginal & $+47.15\pm12.69$ $[23.05,72.05]$ & $+0.03\pm5.15$ $[-10.06,9.83]$\\
Permuted & $-4.59\pm2.14$ $[-9.02,-0.73]$ & $+51.78\pm13.83$ $[26.29,79.34]$\\
Entropy-matched & $+4.38\pm9.88$ $[-15.12,22.88]$ & $+42.81\pm12.40$ $[19.03,66.75]$\\
\midrule
\multicolumn{3}{l}{\emph{IDyOM}}\\
Grammar marginal & $+80.41\pm18.77$ $[49.88,121.63]$ & $+26.61\pm9.76$ $[9.13,46.42]$\\
Permuted & $+0.25\pm2.11$ $[-3.59,4.57]$ & $+106.76\pm25.62$ $[62.61,160.64]$\\
Entropy-matched & $+31.56\pm13.89$ $[6.74,59.82]$ & $+75.45\pm17.29$ $[44.05,110.54]$\\
\bottomrule
\end{tabular}
\end{table}

For the static prior, $b_f$ is the distribution of the first terminal generated from the grammar's predictive start distribution, an equal mixture of the major and minor home chord, computed in closed form through the unary and left-corner closures. Terminal mass is divided equally among the surface symbols that project to it, renormalized and smoothed like $q_E$. The single converged grammar supplies one prior for all stimuli. Both static alternatives retain the event's gate: Static mix is $(1-a_h)q_E+a_hb_f$ and Static tilt is $(1-a_h)q_E+a_hq_Eb_f/\sum_cq_E(c)b_f(c)$. Thus the prior is static but the predictor remains contextual. A sensitivity analysis replaces $b_f$ by the grammar's expected terminal frequency, $p_T\propto\boldsymbol\lambda^\top(I-A)^{-1}E$, with $\boldsymbol\lambda$ the start distribution, $A$ the expected-offspring matrix over categories and $E$ the category-to-terminal emission probabilities; it is finite because the spectral radius of $A$ is $0.960$. In this variant Static tilt reweights $q_E(c)$ by $p_T$ of the terminal of $c$ directly, as Grammar marginal does with the prefix-dependent law, so the tilt does not depend on how many surface symbols share a terminal; Static mix still divides $p_T$ among them. Tonal covariates add tonic-root and dominant-root indicators and $-\log_2b_f$ of the realized chord identically to the Baseline and Ours response models.

The auxiliary whole-song model uses an intercept and each arm's own IC slope, reducing covariance estimation with 29 training songs. Its scores do not replace primary response-LOO (Table~\ref{tab:music-whole-song}). Ours improves over Baseline for every sequence model. Static mix scores below Baseline and Static tilt above Baseline under both priors, but each interval excludes zero only under $p_T$; Ours is distinguished from Static tilt only under $p_T$. Tonal covariates reduce Ours~$-$~Baseline from $+50.27$ to $+22.22$, with an interval that still excludes zero.

\begin{table}[ht]
\centering\small
\caption{Auxiliary whole-song holdout: 29 training songs predict the held-out song, scores averaged over the three sampling streams ($\Delta$ELPD $\pm$ song-cluster SE, 95\% paired song-bootstrap interval). LSTM unless stated.}
\label{tab:music-whole-song}
\begin{tabular}{lrr}
\toprule
Contrast & $\Delta$ELPD $\pm$ SE & 95\% interval\\
\midrule
Ours $-$ Baseline, LSTM & $+50.27\pm12.09$ & $[27.03,74.00]$\\
Ours $-$ Baseline, Transformer & $+38.46\pm11.37$ & $[17.30,61.00]$\\
Ours $-$ Baseline, IDyOM & $+77.78\pm18.32$ & $[45.06,116.09]$\\
Grammar marginal $-$ Baseline & $+38.46\pm10.36$ & $[18.90,58.68]$\\
Ours $-$ Grammar marginal & $+11.81\pm6.48$ & $[0.68,25.46]$\\
Grammar marginal $-$ Baseline, Transformer & $+34.11\pm10.20$ & $[14.29,54.17]$\\
Ours $-$ Grammar marginal, Transformer & $+4.34\pm4.03$ & $[-3.40,12.11]$\\
Grammar marginal $-$ Baseline, IDyOM & $+58.37\pm13.31$ & $[35.25,86.57]$\\
Ours $-$ Grammar marginal, IDyOM & $+19.41\pm6.97$ & $[6.90,33.70]$\\
\midrule
\multicolumn{3}{l}{\emph{Primary static prior $b_f$ (first terminal)}}\\
Static mix $-$ Baseline & $-24.22\pm15.52$ & $[-54.38,5.32]$\\
Static tilt $-$ Baseline & $+28.68\pm15.67$ & $[-1.26,59.64]$\\
Ours $-$ Static mix & $+74.49\pm15.89$ & $[46.52,107.43]$\\
Ours $-$ Static tilt & $+21.59\pm15.67$ & $[-7.12,53.37]$\\
\multicolumn{3}{l}{\emph{Sensitivity prior $p_T$ (expected terminal frequency)}}\\
Static mix $-$ Baseline & $-26.80\pm13.20$ & $[-52.20,-1.11]$\\
Static tilt $-$ Baseline & $+30.21\pm11.22$ & $[9.64,52.84]$\\
Ours $-$ Static mix & $+77.07\pm15.29$ & $[49.18,107.87]$\\
Ours $-$ Static tilt & $+20.06\pm10.23$ & $[0.90,40.70]$\\
\midrule
Ours $-$ Baseline with tonal covariates & $+22.22\pm9.24$ & $[4.77,40.63]$\\
\bottomrule
\end{tabular}
\end{table}

\begin{figure}[t]
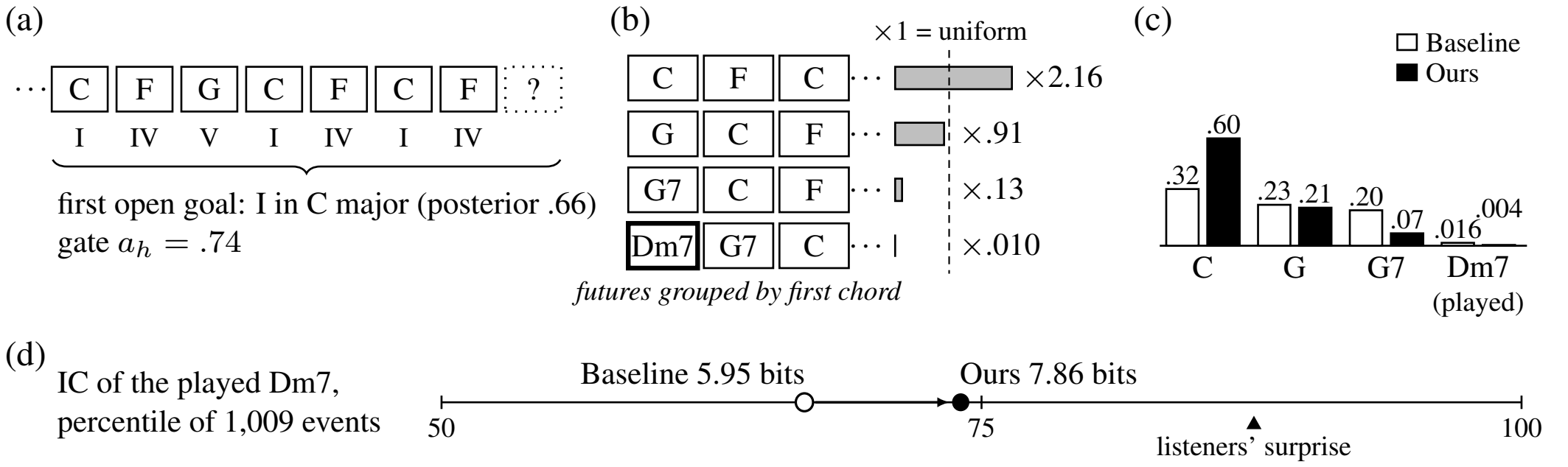

\centering
\workedexampleplot
\caption{\textbf{Worked example} (LSTM, song 22, position 24). (a) The last seven chords of the prefix, the first open goal and the gate. (b) Sampled futures of one stream, grouped by first chord: mean weight relative to uniform, with each group's most frequent opening; the played chord is outlined. (c) Next-chord probability of the three chords most probable under Baseline and of the played chord, averaged over the three streams. (d) Percentile, among all 1,009 events, of the played chord's IC under Baseline and Ours and of listeners' mean surprise.}
\label{fig:worked-example}
\end{figure}

\paragraph{Worked and failure examples.}
\label{app:figure-example}
Figure~\ref{fig:worked-example} traces one event through the LSTM computation: position 24 of song 22 (C major), where the progression had moved mainly between I and IV before listeners heard Dm7. The first open goal is I in C major (posterior $0.66$), and the gate is $a_h=0.74$. Of the 16,384 futures in one stream, 5,491 begin with C, which can complete this goal at once; they receive $2.16$ times the uniform weight on average, against $0.91$ for futures beginning with G, $0.13$ for G7 and $0.010$ for Dm7, which cannot complete the goal in one chord and has low completion probability at longer lengths under the learned grammar. Ours therefore raises the probability of C from $0.32$ to $0.60$ and lowers that of Dm7 from $0.016$ to $0.004$. The IC of the played Dm7 rises from $5.95$ to $7.86$ bits, from the 67th to the 74th percentile of all events, toward the 88th percentile of listeners' surprise.

The largest failure, selected after observing outcomes as the event with the smallest Ours~$-$~Baseline contribution ($-2.35$ nats over both group means), is position 33 of song 27 (A$\flat$ major), where E$\flat$7 has resolved to A$\flat$ and listeners heard D$\flat$ minor. The first open goal is I in E$\flat$ major (posterior $0.39$), and Ours raises the IC of D$\flat$ minor from $9.66$ to $10.71$ bits (85th to 86th percentile), whereas listeners placed it at the 62nd percentile of surprise. Ours~$-$~Baseline is negative at 350 of the 1,009 events.

\subsection{Language: horizon, gate, and comparator sensitivity}
\label{app:language-results}

\begin{figure}[t]
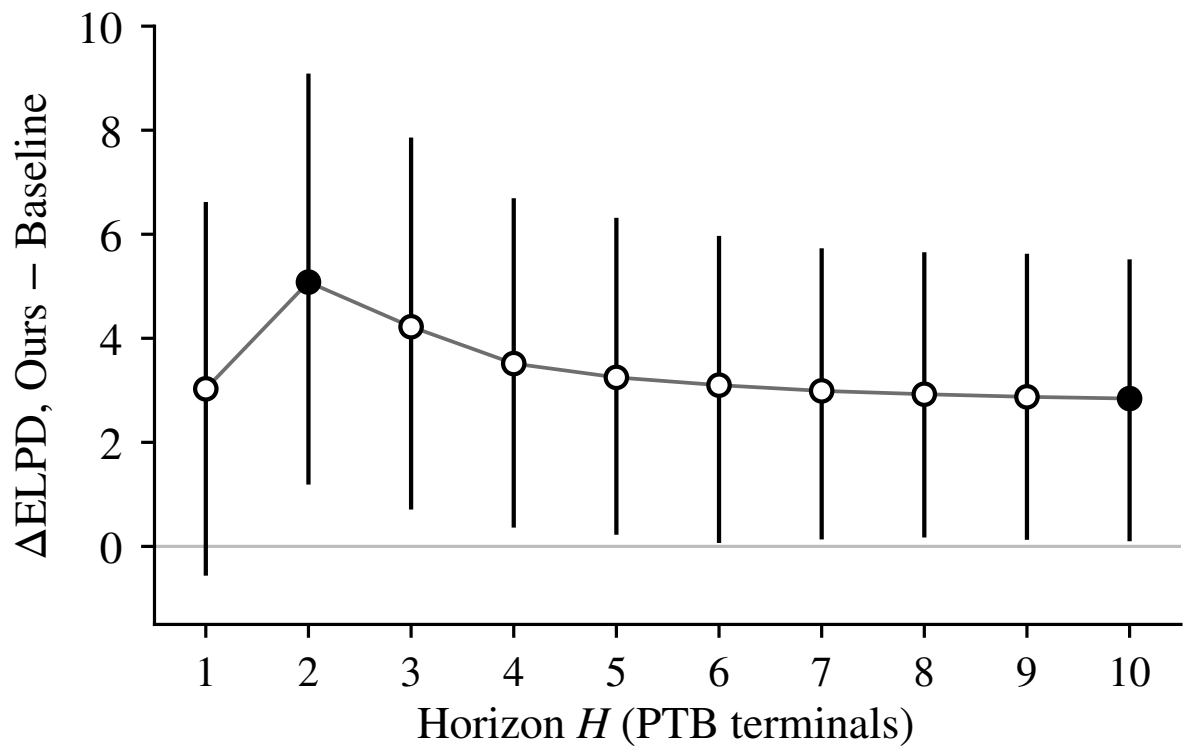

\centering
\languagehorizonplot
\caption{\textbf{Language horizon sensitivity.} Sentence-held-out rating
prediction across horizons measured in PTB terminals.  Points are Ours~$-$~Baseline with 95\% paired sentence-bootstrap intervals; the selected $H=2$ and
the $H=10$ reference are marked. This same-resource sensitivity is not an
estimate of a human planning horizon.}
\label{fig:language-horizon}
\end{figure}

Horizon $H$ truncates a completion-length mixture in PTB terminals, not a fixed word count. The ten-point sweep peaks at $H=2$ (Figure~\ref{fig:language-horizon}); five sentence-grouped inner folds select it from 27 candidate duration rules in every outer fold. Ours ($H=2$) exceeds Ours ($H=1$) by $2.052\pm0.999$ ELPD, interval $[0.086,3.955]$, but the $H=1$/$H=2$ IC correlation is $.99988$ and their joint increment is unresolved, with interval $[-4.913,14.468]$. Nesting protects outer sentences during selection but does not remove dataset-level development.

\begin{table}[t]
\centering\small
\caption{Language rating sensitivity. Intervals resample sentences; PSG, phrase-structure grammar. The upper row gives the $H=10$ reference; the lower block asks whether adding a scalar syntax readout to Baseline is sufficient.}
\label{tab:language-sensitivity}
\begin{tabular}{lrr}
\toprule
Contrast & $\Delta$ELPD $\pm$ SE & 95\% interval\\
\midrule
\multicolumn{3}{l}{\emph{Language ratings: Ours minus Baseline}}\\
$H=10$ reference & $2.842\pm1.392$ & $[0.099,5.517]$\\
\midrule
\multicolumn{3}{l}{\emph{Language ratings: additive syntax predictors, (Baseline + predictor) minus Baseline}}\\
Released PSG surprisal of tags & $0.312\pm1.927$ & $[-3.415,4.161]$\\
PCFG next-terminal surprisal & $-3.489\pm2.106$ & $[-7.747,0.715]$\\
PCFG most-likely-tag probability & $-1.056\pm1.532$ & $[-4.047,1.937]$\\
PCFG tag agreement & $-1.276\pm1.045$ & $[-3.372,0.747]$\\
\bottomrule
\end{tabular}
\end{table}

Compared directly with the corresponding additive models (Table~\ref{tab:language-sensitivity}), Ours gains
$4.768\pm2.356$ ELPD over Baseline plus released PSG tag surprisal,
$8.569\pm3.004$ over Baseline plus PCFG next-terminal surprisal,
$6.136\pm2.657$ over Baseline plus most-likely-tag probability, and
$6.356\pm2.367$ over Baseline plus tag agreement. These comparisons establish the
reported result only for the four tested scalar readouts.

Exact PCFG controls score the first PTB terminal; soft-POS compatibility is a potential rather than a normalized word probability. IAS comparisons use released Euclidean scores, not the unavailable article-final \texttt{cosine\_std} values. The released IAS score adds $9.574\pm4.509$ ELPD to Baseline, and Ours adds $3.671\pm4.910$ after it. Under training-only nested IAS selection these values are $-6.620\pm6.262$ and $+5.336\pm4.986$, respectively. Nested selection uses Gaussian OLS, whereas outer scoring uses sentence random slopes and AR residuals; its loss relative to fixed IAS concentrates in the three folds choosing another configuration. Both Ours-after-IAS increments remain unresolved.

\subsection{Secondary outcomes}
\label{app:language-followups}

For pleasure ratings from 39 participants on the same stimuli \citep{cheungEtAl2024expectancy}, Ours did not reliably improve on Baseline for any of the three sequence models under conditional response-LOO.

In the music EEG, acoustic features alone give a mean held-out correlation of $0.0864$, and adding chord IC and entropy lowers it slightly. For the root-and-quality marginal, Baseline minus acoustic is $-0.00039$ (interval $[-0.00078,-0.00010]$) for the Billboard LSTM and $-0.00055$ ($[-0.00099,-0.00021]$) for the Bach-adapted LSTM, Ours minus acoustic is $-0.00038$ and $-0.00048$, and Ours minus Baseline is $+0.000005$ ($[-0.00011,0.00014]$) and $+0.00007$ ($[-0.00013,0.00035]$). Across both sequence models, three marginals, two annotation variants, participant and piece units, and both $r$ and predictive $R^2$, no chord arm exceeds acoustic features alone, and none of the 48 Ours~$-$~Baseline contrasts survives Holm correction (smallest $p=0.051$). For language N400, Ours minus Baseline is $1.942\pm0.512$ at $H=2$, interval $[0.915,2.907]$. At the $H=10$ reference it is $0.687\pm0.350$, interval $[0.006,1.363]$, and $0.673\pm0.350$, interval $[-0.007,1.352]$, when the gate threshold is reselected within training sentences.

Ours ($H=2$) minus Baseline on log self-paced/first-pass reading times is $0.094\pm0.122$/$-0.239\pm0.278$; adding exact previous-word IC gives $-0.256\pm0.319$/$-0.321\pm0.493$. All four sentence-bootstrap intervals cross zero. Positive observed times use the rating folds and response family; spillover also controls previous-word lexical properties and requires an exact adjacent predecessor. Cloze counts are reconstructed from released responses and correction logs. The target-versus-rest score $k\log p+(n-k)\log(1-p)$ gives Ours a positive but unresolved gain of $+128.61\pm71.66$ nats (sentence-bootstrap interval $[-9.38,271.42]$, 135,923 responses); it does not evaluate the full cloze distribution. Sentence clustering does not capture dependence across items from shared respondents.

\section{Additional music controls}
\label{app:postreview-music}

These analyses use the 30 songs, 1,009 chord events and 2,018 group-mean ratings described in Section~\ref{sec:setup}. Each condition supplies a different surprisal predictor to the same response-model specification, fitted separately and evaluated by conditional response-LOO. All chord predictions use the same smoothing. For conditions based on sampled continuations, we average the held-out rating log scores across three sampling streams before computing uncertainty by paired resampling of songs.

\subsection{Corpus path rescoring}
\label{app:postreview-ngram}

Can simple chord frequencies or chord transitions replace the goal-completion score for a whole continuation? We replace $G_\ell$ in Eq.~\ref{eq:conditional-value} with a product of corpus probabilities for the first $\ell$ chords of a sequence-model continuation $\omega$. The unigram $u$ gives the probability of a chord, and the bigram $b$ gives the probability of a chord conditional on the preceding chord. Both use the same 739 chord symbols as the sequence models and are estimated from the 709 non-stimulus songs (estimation details in Appendix~\ref{app:postreview-simple}). We use the products directly as scores:
\[
S_\ell^{(1)}(\omega,h)=\prod_{j=1}^{\ell}u(\omega_j),\qquad
S_\ell^{(2)}(\omega,h)=b(\omega_1\mid h_{\rm last})\prod_{j=2}^{\ell}b(\omega_j\mid\omega_{j-1}).
\]
The bigram product includes the transition from the last heard chord $h_{\rm last}$ to the first chord of the continuation. As in Ours, we normalize the scores at each length under the same sequence-model continuation distribution and combine the resulting next-chord distributions using the grammar's completion-length weights $d_h(\ell)$. We use the same gate and smoothing, so the comparison tests the score assigned to a continuation while retaining grammar-derived timing and mixing strength. We estimate the distributions for the goal, unigram and bigram path scores using importance-corrected sequential Monte Carlo. We check numerical precision with independent-run resampling and larger sampling budgets, requiring a surprisal interval half-width and a change with increased budget of at most 0.05 bits.

Table~\ref{tab:postreview-ngram} reports the sequential Monte Carlo estimates and a sensitivity analysis that uses the $N=16{,}384$ self-normalized importance-sampling estimates of Ours described in Appendix~\ref{app:probability}, with the n-gram predictions held fixed. The adjusted neural-model intervals use Bonferroni-corrected bootstrap quantiles with a conservative family size of four, separately for unigram and bigram scores. IDyOM comparisons are descriptive.

The neural bigram contrasts are positive after family adjustment. The LSTM unigram contrast is also positive after its separate correction, while the Transformer unigram contrast is uncertain after correction; IDyOM's unigram contrast is uncertain even before correction. The simpler next-chord alternatives are tested separately below.

\begin{table}[H]
\centering\footnotesize\setlength{\tabcolsep}{4pt}
\ifdefined\postreviewblue\color{blue}\fi
\caption{Ours minus corpus path rescoring with the completion-length mixture of Eq.~\ref{eq:conditional-value}. Entries give $\Delta$ELPD with song-cluster SE and 95\% paired song-bootstrap intervals. Adjusted neural-model intervals use Bonferroni correction with a conservative family size of four, separately for each path score. IDyOM and importance-sampling sensitivity rows are descriptive.}
\label{tab:postreview-ngram}
\begin{tabular}{@{}llrll@{}}
\toprule
Model & Path score & $\Delta$ELPD $\pm$ SE & 95\% interval & Adjusted interval\\
\midrule
\multicolumn{5}{l}{\emph{Sequential Monte Carlo estimates}}\\
LSTM & Unigram paths & $+44.03\pm18.35$ & $[9.11,79.78]$ & $[0.14,89.65]$\\
LSTM & Bigram paths & $+56.38\pm18.11$ & $[22.93,92.89]$ & $[14.68,102.61]$\\
Transformer & Unigram paths & $+28.78\pm14.43$ & $[1.84,57.48]$ & $[-5.72,65.49]$\\
Transformer & Bigram paths & $+35.93\pm9.78$ & $[16.98,54.89]$ & $[12.17,60.00]$\\
IDyOM & Unigram paths & $+26.10\pm16.23$ & $[-6.07,56.61]$ & ---\\
IDyOM & Bigram paths & $+60.34\pm22.50$ & $[23.17,107.53]$ & ---\\
\midrule
\multicolumn{5}{l}{\emph{Self-normalized importance-sampling sensitivity}}\\
LSTM & Unigram paths & $+45.49\pm18.90$ & $[10.23,82.73]$ & ---\\
LSTM & Bigram paths & $+57.85\pm18.79$ & $[23.38,96.08]$ & ---\\
Transformer & Unigram paths & $+29.29\pm14.36$ & $[2.41,57.70]$ & ---\\
Transformer & Bigram paths & $+36.44\pm9.69$ & $[17.77,55.23]$ & ---\\
IDyOM & Unigram paths & $+22.27\pm15.96$ & $[-9.30,52.13]$ & ---\\
IDyOM & Bigram paths & $+56.52\pm22.66$ & $[19.22,104.21]$ & ---\\
\bottomrule
\end{tabular}
\end{table}

\subsection{Statistical next-chord corrections}
\label{app:postreview-simple}

These controls ask whether simpler preferences applied directly to the next chord predict ratings as well as Ours. Let $r_h$ be a next-chord distribution. Mix uses $r_h$ as the correction distribution and combines it with $q_E$ as $(1-a_h)q_E+a_hr_h$. Tilt first multiplies $q_E$ by $r_h$ and normalizes the product, then combines it with $q_E$ as $(1-a_h)q_E+a_hq_Er_h/\sum_cq_E(c)r_h(c)$. Both apply the shared smoothing. In Bigram tilt, for example, $r_h(c)=b(c\mid h_{\rm last})$ favors chords that commonly follow the last heard chord.

The unigram and bigram controls use the same corpus distributions as the path scores above. We collapse adjacent identical chord symbols within each of the 709 non-stimulus songs, leaving 70,305 events and 69,596 within-song transitions. The unigram has total Dirichlet mass one with a uniform base. The bigram distributions back off to that unigram with concentration $\tau=100$, selected from $\{1,10,100,1000\}$ by five-fold validation on corpus songs, using chord prediction.

The two static tilts use context-independent chord preferences derived from the grammar: its first-terminal prior $b_f$ and its expected-frequency prior $p_T$ (Appendix~\ref{app:mechanism}). All controls retain the same grammar-derived gate. We fit one response model per deterministic control and sequence model, and compare its pointwise scores with those of Ours averaged across three streams. Tables~\ref{tab:postreview-simple-ours} and~\ref{tab:postreview-simple-baseline} report Ours minus each control and each control minus Baseline. We adjust the intervals for the Static tilt ($b_f$) and Bigram tilt comparisons with Ours separately across the three sequence models.

Ours exceeds Bigram tilt in all three models after family adjustment. For the neural models its differences from either static tilt and from Unigram tilt remain uncertain. For IDyOM the primary Static tilt contrast is positive after adjustment, but the Unigram tilt contrast is uncertain.

\begin{table}[H]
\centering\footnotesize\setlength{\tabcolsep}{4pt}
\ifdefined\postreviewblue\color{blue}\fi
\caption{Ours minus next-chord correction controls under conditional response-LOO. Each cell gives $\Delta$ELPD with song-cluster SE and its 95\% paired song-bootstrap interval. Starred third lines are simultaneous 95\% intervals within each three-model primary family (Static tilt and Bigram tilt).}
\label{tab:postreview-simple-ours}
\begin{tabular}{@{}lccc@{}}
\toprule
Correction & LSTM & Transformer & IDyOM\\
\midrule
Static tilt & \shortstack{$+26.09\pm21.79$\\$[-14.36,70.79]$\\$[-22.63,82.73]^{*}$} & \shortstack{$+14.80\pm14.87$\\$[-13.51,43.63]$\\$[-19.81,50.46]^{*}$} & \shortstack{$+49.81\pm19.10$\\$[15.52,88.95]$\\$[8.36,98.13]^{*}$}\\[3pt]
Static tilt ($p_T$) & \shortstack{$+24.63\pm14.95$\\$[-3.77,55.07]$} & \shortstack{$+17.27\pm11.43$\\$[-4.78,39.66]$} & \shortstack{$+50.51\pm14.69$\\$[24.34,80.89]$}\\[3pt]
Unigram mix & \shortstack{$+77.17\pm35.60$\\$[8.08,147.17]$} & \shortstack{$+89.04\pm29.53$\\$[33.62,148.78]$} & \shortstack{$+67.01\pm27.06$\\$[16.07,120.62]$}\\[3pt]
Unigram tilt & \shortstack{$+38.02\pm20.12$\\$[-0.29,78.17]$} & \shortstack{$+22.68\pm11.90$\\$[-0.18,45.95]$} & \shortstack{$+24.74\pm16.68$\\$[-8.51,56.53]$}\\[3pt]
Bigram mix & \shortstack{$+68.23\pm32.76$\\$[7.18,134.58]$} & \shortstack{$+79.18\pm26.46$\\$[29.81,131.25]$} & \shortstack{$+55.77\pm24.74$\\$[9.46,104.98]$}\\[3pt]
Bigram tilt & \shortstack{$+56.70\pm21.76$\\$[17.11,102.18]$\\$[10.38,114.15]^{*}$} & \shortstack{$+34.19\pm10.62$\\$[13.63,55.20]$\\$[8.89,59.88]^{*}$} & \shortstack{$+61.68\pm24.68$\\$[20.74,115.48]$\\$[13.94,129.22]^{*}$}\\[3pt]
\bottomrule
\end{tabular}
\end{table}

\begin{table}[H]
\centering\footnotesize\setlength{\tabcolsep}{4pt}
\ifdefined\postreviewblue\color{blue}\fi
\caption{Next-chord correction controls minus Baseline under conditional response-LOO. Each cell gives $\Delta$ELPD with song-cluster SE and its 95\% paired song-bootstrap interval.}
\label{tab:postreview-simple-baseline}
\begin{tabular}{@{}lccc@{}}
\toprule
Correction & LSTM & Transformer & IDyOM\\
\midrule
Static tilt & \shortstack{$+35.65\pm21.47$\\$[-4.67,78.89]$} & \shortstack{$+32.39\pm17.73$\\$[-2.56,66.25]$} & \shortstack{$+57.20\pm17.07$\\$[24.00,90.51]$}\\[3pt]
Static tilt ($p_T$) & \shortstack{$+37.11\pm15.29$\\$[9.17,68.73]$} & \shortstack{$+29.91\pm12.91$\\$[4.74,54.99]$} & \shortstack{$+56.51\pm16.50$\\$[26.89,91.21]$}\\[3pt]
Unigram mix & \shortstack{$-15.42\pm31.27$\\$[-74.31,46.16]$} & \shortstack{$-41.85\pm27.13$\\$[-98.63,7.38]$} & \shortstack{$+40.00\pm38.79$\\$[-29.69,120.22]$}\\[3pt]
Unigram tilt & \shortstack{$+23.72\pm16.78$\\$[-7.58,57.28]$} & \shortstack{$+24.51\pm15.09$\\$[-4.89,54.11]$} & \shortstack{$+82.27\pm27.64$\\$[35.14,140.98]$}\\[3pt]
Bigram mix & \shortstack{$-6.48\pm27.51$\\$[-60.93,45.14]$} & \shortstack{$-32.00\pm20.81$\\$[-73.62,7.13]$} & \shortstack{$+51.24\pm35.23$\\$[-11.20,124.45]$}\\[3pt]
Bigram tilt & \shortstack{$+5.04\pm16.35$\\$[-28.56,34.58]$} & \shortstack{$+13.00\pm12.89$\\$[-11.60,38.54]$} & \shortstack{$+45.34\pm16.27$\\$[18.01,80.52]$}\\[3pt]
\bottomrule
\end{tabular}
\end{table}

\subsection{Learned grammar probabilities}
\label{app:postreview-weights}

Do the grammar probabilities learned from the corpus improve rating prediction over uniform or random probabilities on the same grammar rules? For this comparison we use the LSTM and hold the rule inventory, 168 categories, chord representation and sampled continuations fixed. Uniform assigns equal probability to every allowed action for each scale degree and mode, including the ten chord-form emissions. This is the mean under the Dirichlet$(1)$ prior on each row of probabilities. Random uses three independent draws from this prior (101, 202 and 303). Changing the probabilities changes the grammar's analysis of the prefix, completion scores, length weights and entropy gate. We also repeat the comparison with a common constant gate of $0.620$ to test whether learning helps when every grammar has the same mixture weight.

The learned grammar exceeds both Uniform and the mean random score with its own gate and with the common constant gate (Table~\ref{tab:postreview-weights}). The random mean averages pointwise response log scores over the three grammars and proposal streams; each individual draw is also shown. Learning changes structural and chord-emission preferences as well as induced goals and lengths, so the contrast measures their combined effect.

\begin{table}[H]
\centering\footnotesize\setlength{\tabcolsep}{4pt}
\ifdefined\postreviewblue\color{blue}\fi
\caption{LSTM grammar-probability controls: $\Delta$ELPD with song-cluster SE and 95\% paired song-bootstrap intervals. Intervals are unadjusted and conditional on the three specified random grammars; the sampling unit is the song.}
\label{tab:postreview-weights}
\begin{tabular}{@{}lcc@{}}
\toprule
Comparison & Own entropy gate & Common constant gate\\
\midrule
Learned $-$ Baseline & \shortstack{$+61.74\pm16.14$\\$[30.52,93.67]$} & \shortstack{$+59.95\pm15.79$\\$[28.74,90.82]$}\\[2pt]
Uniform $-$ Baseline & \shortstack{$+19.05\pm17.34$\\$[-14.79,52.59]$} & \shortstack{$+20.26\pm15.24$\\$[-9.13,50.34]$}\\[2pt]
Random 101 $-$ Baseline & \shortstack{$-13.43\pm23.83$\\$[-61.57,29.97]$} & \shortstack{$-2.33\pm18.79$\\$[-39.78,32.36]$}\\[2pt]
Random 202 $-$ Baseline & \shortstack{$-7.59\pm12.90$\\$[-32.61,17.75]$} & \shortstack{$+0.95\pm12.00$\\$[-21.76,25.22]$}\\[2pt]
Random 303 $-$ Baseline & \shortstack{$+3.78\pm20.21$\\$[-36.11,42.45]$} & \shortstack{$+11.41\pm15.64$\\$[-19.31,40.95]$}\\[2pt]
Random mean $-$ Baseline & \shortstack{$-5.75\pm16.00$\\$[-36.72,24.73]$} & \shortstack{$+3.34\pm12.44$\\$[-20.40,27.66]$}\\[2pt]
Learned $-$ Uniform & \shortstack{$+42.69\pm18.23$\\$[9.05,79.29]$} & \shortstack{$+39.69\pm17.08$\\$[7.81,73.54]$}\\[2pt]
Learned $-$ Random 101 & \shortstack{$+75.17\pm26.06$\\$[28.12,127.33]$} & \shortstack{$+62.28\pm22.95$\\$[19.85,107.84]$}\\[2pt]
Learned $-$ Random 202 & \shortstack{$+69.33\pm16.07$\\$[39.75,101.83]$} & \shortstack{$+59.00\pm14.89$\\$[31.10,88.88]$}\\[2pt]
Learned $-$ Random 303 & \shortstack{$+57.96\pm21.20$\\$[18.80,100.41]$} & \shortstack{$+48.54\pm19.34$\\$[13.53,87.41]$}\\[2pt]
Learned $-$ Random mean & \shortstack{$+67.49\pm18.37$\\$[33.14,103.69]$} & \shortstack{$+56.61\pm16.78$\\$[25.13,89.49]$}\\[2pt]
\bottomrule
\end{tabular}
\end{table}

\subsection{Matching the completion-length mixture}
\label{app:postreview-length}

In Section~\ref{sec:results-mechanism}, Grammar marginal sums over all possible goal-completion lengths. Ours instead limits the lengths it considers and normalizes and mixes the predictions separately by length. We construct two controls to test whether these differences account for the rating-prediction gap between the two methods. Let $A_h(\ell,t)$ be the grammar probability that the tracked goals yield exactly $\ell$ terminals, starting with terminal $t$, averaged over the same prefix posterior. With $v(c)$ mapping a literal chord to its grammar terminal, define the ungated distributions
\[
\begin{aligned}
T(c\mid h)&\propto q_E(c\mid h)\sum_{\ell=1}^{H}A_h(\ell,v(c)),\\
M(c\mid h)&=\sum_{\ell=1}^{H}d_h(\ell)
\frac{q_E(c\mid h)A_h(\ell,v(c))}{\sum_{c'}q_E(c'\mid h)A_h(\ell,v(c'))}.
\end{aligned}
\]
Both controls use $H=10$ and the same gate and smoothing as Ours. $T$ applies the length limit to Grammar marginal. $M$ additionally matches how Ours normalizes at each length and combines the predictions with $d_h(\ell)$. The comparison with $M$ asks whether scoring the sequence model's continuations improves rating prediction when the length limit and mixture are shared. We compute the grammar masses by exact inside recursions. Ours uses the self-normalized importance-sampling estimator of Appendix~\ref{app:probability}. Across the six response fits for the deterministic controls, maximum $\hat R$ is $1.00388$, minimum parameter ESS is $1564.9$, maximum Pareto $k$ is $0.3771$ and maximum pointwise MCSE is $0.00547$.

The IDyOM Ours--$M$ increment remains positive after adjustment across three models, while the LSTM and Transformer increments remain uncertain (Table~\ref{tab:postreview-length}). All six secondary decomposition intervals include zero after adjustment. Intervals use 20,000 paired bootstrap samples of the 30 songs.

\begin{table}[H]
\centering\footnotesize\setlength{\tabcolsep}{4pt}
\ifdefined\postreviewblue\color{blue}\fi
\caption{Length matching under conditional response-LOO: $\Delta$ELPD with song-cluster SE, ordinary 95\% paired song-bootstrap interval and family-adjusted interval. The three Ours--$M$ contrasts form the primary family; six adjacent decomposition contrasts form a secondary family. Table~\ref{tab:music-controls} reports the comparison of Ours with Grammar marginal.}
\label{tab:postreview-length}
\begin{tabular}{@{}lrllr@{}}
\toprule
Comparison & $\Delta$ELPD $\pm$ SE & 95\% interval & Adjusted interval & Family size\\
\midrule
\multicolumn{5}{l}{\emph{LSTM}}\\
Ours $-$ $M$ & $+7.36\pm7.21$ & $[-5.16,22.30]$ & $[-7.39,26.23]$ & 3\\
$T$ $-$ Grammar marginal & $+0.66\pm0.87$ & $[-1.05,2.35]$ & $[-1.65,2.90]$ & 6\\
$M$ $-$ $T$ & $+2.01\pm1.63$ & $[-0.94,5.28]$ & $[-1.84,6.56]$ & 6\\
\midrule
\multicolumn{5}{l}{\emph{Transformer}}\\
Ours $-$ $M$ & $-1.16\pm4.96$ & $[-10.96,8.35]$ & $[-13.36,10.50]$ & 3\\
$T$ $-$ Grammar marginal & $+1.50\pm0.86$ & $[-0.10,3.19]$ & $[-0.62,3.78]$ & 6\\
$M$ $-$ $T$ & $-0.30\pm1.56$ & $[-3.26,2.70]$ & $[-4.27,3.71]$ & 6\\
\midrule
\multicolumn{5}{l}{\emph{IDyOM}}\\
Ours $-$ $M$ & $+24.04\pm9.45$ & $[6.86,43.27]$ & $[3.64,48.57]$ & 3\\
$T$ $-$ Grammar marginal & $+2.35\pm0.97$ & $[0.53,4.28]$ & $[-0.09,4.95]$ & 6\\
$M$ $-$ $T$ & $+0.21\pm1.87$ & $[-3.06,4.13]$ & $[-4.03,5.65]$ & 6\\
\bottomrule
\end{tabular}
\end{table}

\section{Numerical diagnostics}
\label{app:reproducibility}

The goal-certainty gate has mean $0.620$ and range $0.342$--$0.858$ over the 1,009 music events; it depends only on the prefix and is therefore identical for the three sequence models. The effective sample size of our model's importance weights, out of $N=16{,}384$ futures, has median $4{,}117$, $3{,}679$ and $3{,}574$ for the LSTM, Transformer and IDyOM, with minima $18.0$, $44.1$ and $54.4$. Recomputed Baseline IC matches each sequence model's Baseline predictor within $10^{-6}$ bits, and recomputing Ours inside the control pipeline reproduces its probabilities within $5\times10^{-8}$. All 33 surprise response fits (Ours and its Grammar marginal, Permuted and Entropy-matched controls for the three sequence models, and the LSTM Constant gate) have no divergent transitions or maximum-tree-depth hits, $\hat R\le1.005$ and minimum parameter ESS $1{,}669$; their integrated response-LOO scores have maximum Pareto $k=0.361$ and maximum pointwise MCSE $0.005$. The 81 additional horizon fits ($H=1,\dots,9$) have no divergent transitions, $\hat R\le1.006$, maximum Pareto $k=0.356$ and maximum pointwise MCSE $0.005$. Remaining limits include data access, environment portability, approximate computational uncertainty, and inference beyond observed language proposal support.
\label{app:appendix-end}

\end{document}